\pdfoutput=1

\documentclass[11pt]{article}

\usepackage[final]{acl}

\usepackage{times}
\usepackage{latexsym}

\usepackage[T1]{fontenc}

\usepackage[utf8]{inputenc}

\usepackage{microtype}

\usepackage{inconsolata}

\usepackage{graphicx}
\usepackage{subcaption} 
\usepackage{float} 
\usepackage{soul}
\usepackage{amssymb}
\usepackage{amsmath}
\usepackage{tcolorbox}
\usepackage{enumitem} 
\usepackage{booktabs}  
\usepackage{xcolor} 
\usepackage{colortbl}
\usepackage{arydshln}
\usepackage{multirow}
\usepackage{lmodern}
\usepackage{booktabs}

\title{RMoA: Optimizing Mixture-of-Agents through Diversity Maximization and Residual Compensation}

\author{
 \textbf{Zhentao Xie\textsuperscript{1,2}},
 \textbf{Chengcheng Han\textsuperscript{2}},
 \textbf{Jinxin Shi\textsuperscript{1}},
 \textbf{Wenjun Cui\textsuperscript{3}},
\\
 \textbf{Xin Zhao\textsuperscript{4}},
 \textbf{Xingjiao Wu\textsuperscript{1}},
 \textbf{Jiabao Zhao\textsuperscript{3} \thanks{Corresponding author}}
\\
\\
 \textsuperscript{1}East China Normal University
 \textsuperscript{2}Meituan Inc.
 \textsuperscript{3}Donghua University
 \textsuperscript{4}Tsinghua University
\\
 \small{
xiezhentao@stu.ecnu.edu.cn,jbzhao@dhu.edu.cn
 }
}

\begin{document}
\maketitle
\begin{abstract}
Although multi-agent systems based on large language models show strong capabilities on multiple tasks, they are still limited by high computational overhead, information loss, and robustness. Inspired by ResNet’s residual learning, we propose Residual Mixture-of-Agents (RMoA), integrating residual connections to optimize efficiency and reliability. To maximize information utilization from model responses while minimizing computational costs, we innovatively design an embedding-based diversity selection mechanism that greedily selects responses via vector similarity. Furthermore, to mitigate iterative information degradation, we introduce a Residual Extraction Agent to preserve cross-layer incremental information by capturing inter-layer response differences, coupled with a Residual Aggregation Agent for hierarchical information integration. Additionally, we propose an adaptive termination mechanism that dynamically halts processing based on residual convergence, further improving inference efficiency. RMoA achieves state-of-the-art performance on the benchmarks of across alignment, mathematical reasoning, code generation, and multitasking understanding, while significantly reducing computational overhead. Code is available at \url{https://github.com/mindhunter01/RMoA}.

\end{abstract}

\section{Introduction}
Large language models (LLMs) \citep{achiam2023gpt, team2024gemma, yang2024qwen2} have achieved significant advancements in extensive natural language processing tasks \citep{wang2022self,xu2024wizardlm}. 
Recently, researchers have proposed several policy-based methods that enhance model performance without  model scaling. Notable approaches include Chain-of-Thought  \citep{wei2022chain}, which enhances multi-step reasoning; Retrieval-Augmented Generation (RAG)  \citep{lewis2020retrieval}, which leverages external information sources; and Multi-Agent Systems (MAS)  \citep{liang2023encouraging,li2023camel}. Among these innovations, MAS has garnered significant attention due to exceptional flexibility and broad compatibility.

Recently, iterative collaboration strategies have been shown to enhance the capabilities of MAS. \citet{wang2025mixtureofagents} proposed the Mixture-of-Agents (MoA) architecture.
This architecture leverages a hierarchical processor design that enables multiple layers of agents to process queries in parallel, significantly improving computational efficiency. Then, MoA employs an aggregator to integrate the outputs from these agents, generating the final response.
Subsequently, Sparse Mixture-of-Agents (SMoA)  \citep{li2024smoa} were introduced to reduce the large number of tokens involved in parallel queries under MoA, thus lowering inference costs. 
These approaches incorporate a judge model \citep{zheng2023judging} to evaluate the quality of responses generated by different models, thereby reducing the number of tokens processed by the aggregator. While this strategy somewhat alleviates computational overhead, it still faces challenges in ensuring the robustness of quality differentiation among responses \citep{dhurandhar2024ranking}. Moreover, as the number of processing layers increases, MoA may suffer from the loss of critical information during aggregation \citep{tworkowski2024focused}, leading to inaccurate responses and ultimately compromising the overall stability and reliability .

To address these challenges, we propose RMoA, an improved MoA-based architecture inspired by residual connections. Unlike existing approaches, we do not employ a judge model to select the optimal response. Instead, we introduce an embedding model to convert responses into vector representations and compute their similarities. A greedy strategy is then applied to select $K$ responses with the highest diversity, ensuring greater information heterogeneity. Additionally, we design a Residual Extraction Agent to capture differences between responses at successive layers. These residuals, along with the selected diverse responses, are fed into the aggregator, preserving incremental information and mitigating the loss of key content during deep aggregation.

To conserve computational resources, we incorporate an Adaptive Termination Mechanism, which dynamically determines when to halt processing based on response variations between iterations, thereby reducing unnecessary overhead. Furthermore, to foster diverse and creative reasoning, each agent is assigned a distinct role-playing persona.

To comprehensively evaluate the effectiveness of our approach, we conduct extensive experiments on alignment, mathematics, code generation, and multi-task understanding. Experimental results demonstrate that RMoA achieves state-of-the-art performances with lower computational costs. Additionally, a series of ablation studies validate the effectiveness of each component in RMoA. Finally, we investigate RMoA's performance under increased computational budgets, showing that for models with strong general capabilities, deeper architectures tend to yield improved performance across most datasets.

Overall, our contributions consist of three parts.
\begin{itemize}
    \item  We introduce RMoA, an improved MoA architecture with an embedding-based selection mechanism, a Residual Extraction Agent, and an Adaptive Termination Mechanism to enhance efficiency and diversity.
    \item  We validate RMoA on multiple benchmarks, demonstrating superior performance with lower computational cost. Ablation studies confirm the effectiveness of each component.
    \item  We analyze RMoA under varying computational budgets, showing that deeper architectures improve performance and providing insights into scalable multi-agent systems.
\end{itemize}

\section{Related Work}
\subsection{LLM Reasoning}
Recent advancements in LLM reasoning have introduced various prompt strategies to improve downstream tasks. Chain of Thought (CoT) \citep{wei2022chain,kojima2022large} prompting guides the model to explicitly output the intermediate step-by-step reasoning before providing the final answer. To address errors in CoT, such as missing steps or inconsistent logic, Auto-CoT \citep{zhang2022automatic} automates the generation of diverse demonstrations, while Reprompting \citep{xu2023reprompting} iteratively refines prompts to enhance reasoning. Plan-and-Solve (PS) \citep{wang2023plan} Prompting introduces a planning phase to break tasks into subtasks with detailed instructions. Additionally, LogiCoT \citep{liu2023logicot} integrates symbolic logic to validate reasoning processes and reduce errors. Building on the linear structure of CoT, Tree of Thought (ToT) \citep{yao2023tree} expands CoT with a tree-like structure, considering multiple reasoning paths and self-evaluating choices, and Graph of Thought (GoT) \citep{besta2024graph} represents reasoning steps as graph nodes, incorporating operations like aggregation and refinement for complex tasks. Addtionally, Cumulative Reasoning (CR) \citep{zhang2023cumulative} simulates human-like iterative reasoning, while LeMa \citep{an2023learning} uses GPT-4 as an error-correcting agent to revise faulty reasoning steps and fine-tune LLMs. However, the existing topological relationship (such as linear chain or tree structure) is usually fixed in advance , which lacks dynamic adaptability and extensibility.

\subsection{Collaborative Agents}
Collaborative Agents in LLM-based systems enhance task performance by enabling agents to work together, share knowledge, and dynamically adjust their strategies to solve complex problems\citep{guo2024large,du2025survey}. Peer Review Collaboration \citep{Xu2023TowardsRI} refines solutions based on feedback from other agents. The Chain of Experts framework \citep{xiao2023chain} coordinates agents with specialized knowledge to solve complex tasks, while Theory of Mind \citep{li2023theory} improves collaboration by enabling agents to predict each other’s intentions. Besides, frameworks like MetaGPT \citep{hong2023metagpt} and Chatdev \citep{qian2024chatdev} utilize specialized agents for modular tasks, such as programming, while MapCoder \citep{islam2024mapcoder} extends this approach by integrating agents for code retrieval, planning, and debugging. Dynamic frameworks like DyLAN \citep{liu2024dynamic} and MACNET \citep{qian2025scaling} organize agent interactions based on task importance, improving scalability and solution quality in large collaborations. \citet{wang2025mixtureofagents, li2024smoa} proposed Mixture-of-Agents architecture for iterative collaboration. However, they are still limited by high computational overhead, information loss, and robustness. 

\begin{figure}[t]
  \includegraphics[width=\columnwidth]{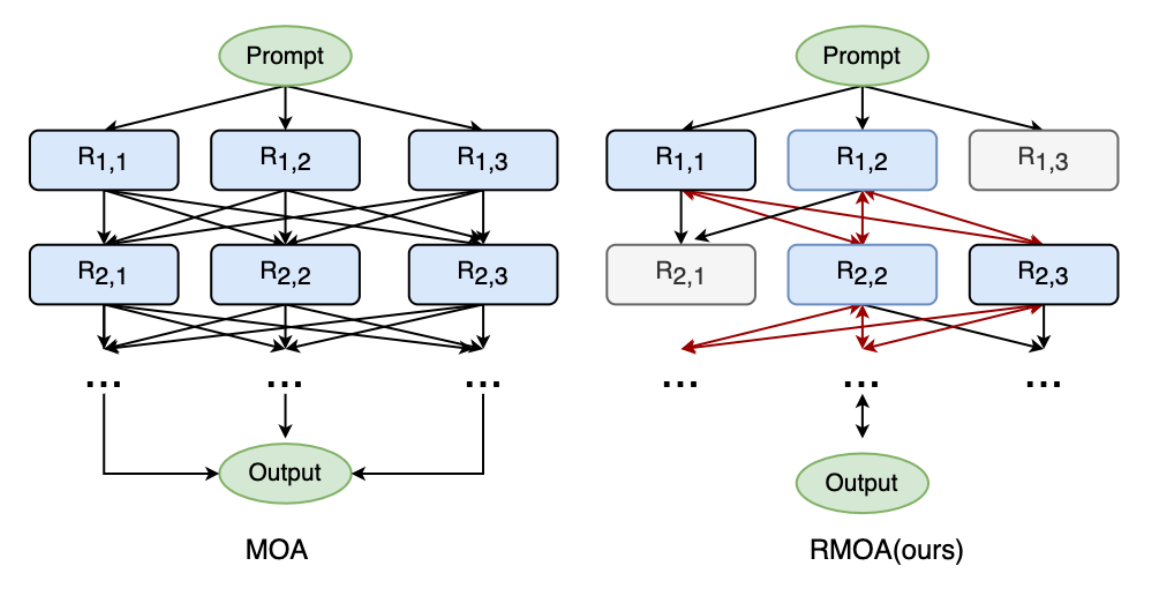}
  \caption{MoA-RMoA Structural Comparison.}
  \label{fig:Figure 1}
\end{figure} 

\begin{figure*}[t]
  \includegraphics[width=\linewidth]{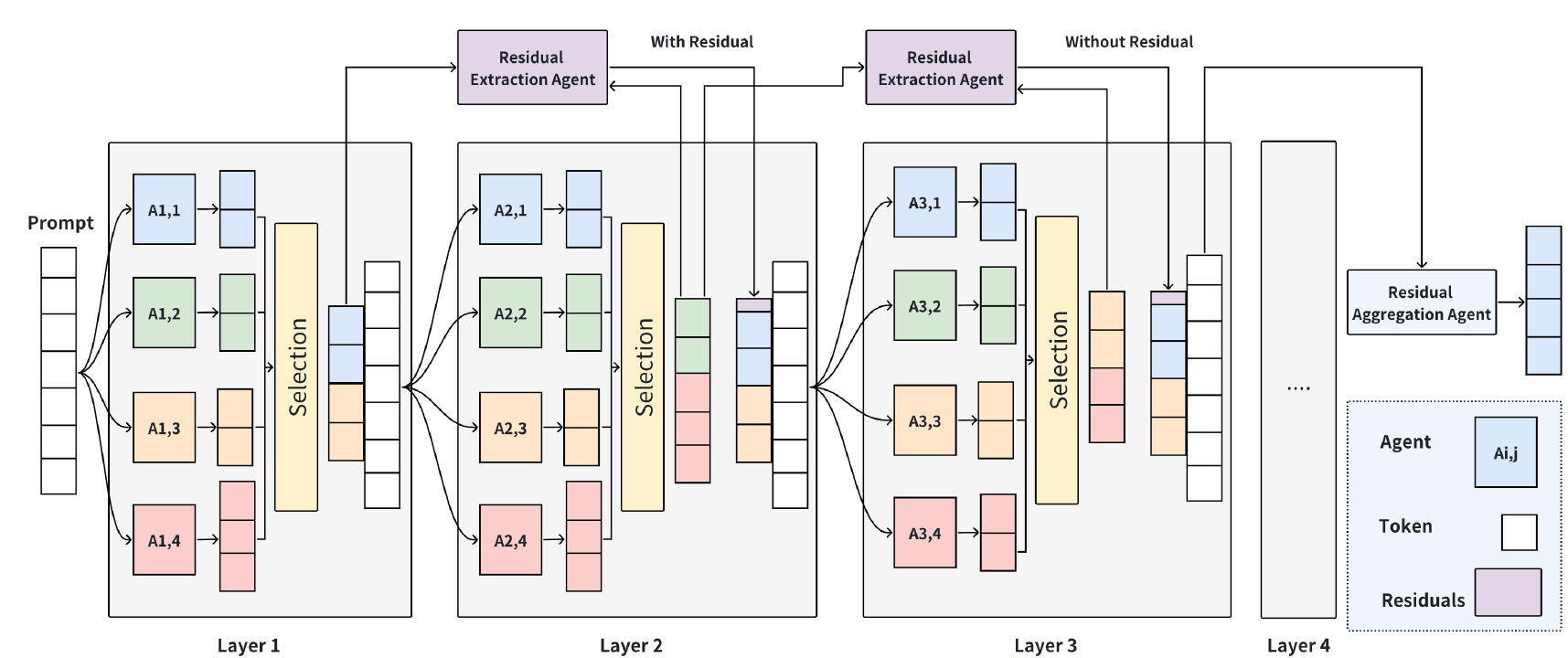}
  \caption{Overview of Residual Mixture-of-Agents Structure.}
  \label{fig:Figure 2}
\end{figure*} 

\section{Methodology}
This section begins with an overview of MoA, followed by a comprehensive analysis of RMoA's core components: Greedy Diversity Embedding Selection, Residual Agent, and Adaptive Termination mechanisms, as depicted in Figure~\ref{fig:Figure 2}. Initially, the Greedy Diversity Embedding Selection filters out diverse and representative responses, ensuring varied inputs for further processing. Next, the Residual Agent uses these selected responses to pinpoint key differences between dialogue rounds, integrating them into the reference material to reduce information loss. Finally, the Adaptive Termination Mechanism continuously monitors the process in real-time, deciding whether to continue based on residual detection outcomes, thus avoiding unnecessary iterations and potential hallucinations.

\subsection{Mixture-of-Agents}
As shown in Figure~\ref{fig:Figure 1}, MoA employs a multi-layered architecture to generate and optimize responses. The structure comprises $ L $ layers, with each layer $ l $ consisting of $ N $ agents, denoted as $\{A_{l,1}, A_{l,2}, \ldots, A_{l,N}\}$. 
And an aggregator $A_g$ is positioned at the final. Initially, in the first layer, multiple proposers independently generate initial responses $\{R_{1,1}, R_{1,2}, ..., R_{1,N}\}$ to a given query $x$. These responses are then concatenated as $R_1$ and serve as input for the subsequent layer. This iterative process continues until reaching the final layer, producing output $R_L$. Finally, all inputs are fed into $A_g$ for integration and optimization, generating the final response $R_F$.

\subsection{Residual Mixture-of Agents}
\subsubsection{Greedy Diversity Embedding Selection}
\citet{tworkowski2024focused} identified the "Distraction issue," where increasing tokens in the self-attention mechanism can cause semantic overlap among keys, hindering the model's focus on relevant information. In the MoA, generating responses by referencing up to $N$ previous models' responses increases cognitive load. To address this, we use a greedy strategy to maximize diversity, selecting $K$ diverse responses for concatenation.

Taking layer $l$ as an example, our objective is to identify a subset containing only $K$ elements from all responses $\{R_{l,1}, R_{l,2}, ..., R_{l,N}\}$ through maximum semantic diversity. The algorithm follows these steps:

 \textbf{In the Similarity Matrix Construction} involves computing the cosine similarity matrix \( S \in \mathbb{R}^{n \times n} \) for all pairs of responses. Each element of this matrix is defined by the cosine similarity between embedding vectors \( e_i \) and \( e_j \):
\begin{equation}
S_{i,j} = \cos(e_i, e_j) = \frac{e_i \cdot e_j}{\|e_i\| \|e_j\|},
\end{equation}
where $e_i$ and $e_j$ represent the embedding vectors of $R_{l,i}$ and $R_{l,j}$, respectively. This matrix is symmetric, meaning that \( S_{i,j} = S_{j,i} \), and the diagonal elements are equal to 1, \( S_{i,i} = 1 \).

\textbf{In the Initialization Phase}, we begin by defining the candidate index set \( C = \{ 1, 2, \ldots, N \} \) and the selected index set \( Q = \emptyset \). The initial element is selected by minimizing the global average similarity, which is calculated as:
\begin{equation}
i_0 = \arg\min_{i \in C} \left( \frac{1}{N} \sum_{j=1}^{N} S_{i,j} \right).
\end{equation}
Once the initial element \( i_0 \) is identified, the sets are updated accordingly: \( Q \leftarrow Q \cup \{ i_0 \} \) and \( C \leftarrow C \setminus \{ i_0 \} \).

\textbf{In the Iterative Selection Phase} involves selecting elements to maximize diversity. For each iteration \( t = 1, \ldots, K-1 \), the process begins with the \textit{Maximum Similarity Calculation}, where for each candidate \( i \in C \), the maximum similarity with the already selected set is computed as follows: 
\begin{equation}
\Phi(i) = \max_{q \in Q} S_{i,q}.
\end{equation}
Following this, the \textit{Minimization Selection} step chooses the candidate that minimizes \( \Phi(i) \):

\begin{equation}
i_t = \arg\min_{i \in C} \Phi(i).
\end{equation}
After selecting the candidate, the sets are updated: \( Q \leftarrow Q \cup \{ i_t \} \) and \( C \leftarrow C \setminus \{ i_t \} \). This iterative process continues until the termination condition \( |Q| = K \) is met, resulting in the final selected reference text set \( S = \{ r_i \mid i \in Q \} \).

\subsubsection{Residual Agent}
In the MoA framework, models may experience information loss when referencing multiple responses from the previous iteration. This phenomenon can lead to a gradual degradation of information, causing the model's performance to decline even in early stages \citep{li2024smoa}.
We draw inspiration from the residual concept in ResNet and introduce a residual extraction agent and a residual aggregation agent.
The residual extraction agent employs predefined prompt templates to identify significant variations between consecutive dialogue responses, integrating these differential features with the previous layer's output to provide contextual input for subsequent processing. At the architecture's final stage, the residual aggregation agent combines the preceding layer's reference response with the current layer's residual features to generate an optimized system output.

\textbf{Residual Extraction Agent}\quad In layer \(l\), following the execution of Greedy Diversity Embedding Selection, we obtain \(K\) candidate responses \(\{R_{l,x_j}\}_{j=1}^{K}\) generated by proposers, along with historical responses \(\{R_{l-1,x_j}\}_{j=1}^{K}\) from the preceding layer \((l-1)\). These responses are concatenated to form composite inputs, which are subsequently processed by the Residual Extraction Agent \(Res\) to identify useful differences. This process can be formally represented as follows:

\begin{itemize}
    \item \textbf{Concatenation Operation:} Concatenate the \( l \)-th layer's responses with the previous layer's aggregated response:
    \begin{equation}
        R_l = \text{Cat}( \{R_{l,x_j}\}_{j=1}^{K}, \{R_{l-1,x_j}\}_{j=1}^{K}).
    \end{equation}
    
    \item \textbf{Residual Extraction:} Compute the residual using the residual extraction agent \( \text{Res} \):
    \begin{equation}
        \Delta R_l = \text{Res}(R_l, prompt).
    \end{equation}
    
    \item \textbf{Residual Reference:} Concatenate the extracted residual \( \Delta R_i \) with the previous layer's responses to provide reference for the next layer:
    \begin{equation}
        \hat{R}_l = \text{Cat}(\{R_{l-1,x_j}\}_{j=1}^{k}, \Delta R_l).
    \end{equation}
\end{itemize}
It is noteworthy that when \( l = 1 \), there is no aggregated response from the previous layer, so we set \( \Delta R_0 \) to be empty.

\textbf{Residual Aggregation Agent} \quad The Residual Aggregation Agent acts on the final layer to integrate the model's responses. Specifically, for the last layer \(l\), after greedy differential embedding selection and residual extraction, the responses obtained from the previous round are aggregated with the current round's residual \(\Delta R_l\):
\begin{equation}
   R_l = \text{Agg}(\{R_{l-1,x_j}\}_{j=1}^{k}, \Delta R_l).
\end{equation}
Through this approach, the RMoA framework effectively captures and integrates differences across iterations, minimizing information loss and enhancing model performance in multi-layer iterative processes.

\subsubsection{Adaptive Termination }
In the MoA framework, typically \( l \) layers of processing are required to obtain the final output of a problem. However, sometimes the ideal result may be achieved at a shallower layer, and continuing the process might lead to unnecessary computation or even negative effects. To address this, we introduce an adaptive stopping mechanism that determines whether to continue iteration by detecting the presence of residuals in the extraction process.

Specifically, the core of the adaptive stopping mechanism is: if no residuals are detected in the current layer and the preceding \( m \) consecutive layers, the iteration process is terminated early. Mathematically, this can be expressed as:

For a given layer \( i \), if for all \( j = 0, 1, \ldots, m-1 \), the values of \( \Delta R_{i-j} \) are "no change" or "no update", then stop the iteration. Otherwise, continue processing to the next layer.

 This mechanism reduces unnecessary computational resource consumption while ensuring result quality, thereby improving the model's efficiency and performance. By adaptively determining the presence of residuals, the model can dynamically adjust the depth of processing, avoiding over-computation and optimizing performance.

\section{Evaluation}
\subsection{Setup}
\textbf{Benchmark}\quad To comprehensively evaluate the effectiveness of our method, we conduct experiments across four critical benchmarks: alignment, mathematical reasoning, general reasoning, and code understanding. For alignment assessment, we employ AlpacaEval 2.0 \citep{dubois2024length} with gpt-4-1106-preview as the reference model. This benchmark utilizes a GPT-4-based evaluator to calculate length-controlled (LC) win rates, effectively mitigating length bias while comparing model responses against the reference outputs.

For mathematical reasoning evaluation, we adopt the MATH \citep{hendrycks2021measuring} benchmark, which contains 5,000 challenging competition-level mathematics problems requiring multi-step reasoning. The general reasoning capability is measured through MMLU-redux \citep{gema2024we}, a refined subset of the MMLU \citep{hendrycks2020measuring} benchmark comprising 3,000 manually re-annotated samples that address original dataset errors while maintaining comprehensive knowledge coverage.CRUX \citep{gu2024cruxeval} is a benchmark for assessing code understanding, featuring 800 Python functions. It evaluates input and output prediction tasks, requiring advanced code comprehension and reasoning.

\textbf{Implementation Details} \quad In our research, we mainly developed RMoA using open-source small models, achieving significant performance improvements across multiple datasets, including Gemma2-9B-Instruct  \citep{team2024gemma}, Qwen2.5-7b-Instruct  \citep{yang2024qwen2}, and Llama3.1-8b-Instruct  \citep{vavekanand2024llama}. We build up to 6 layers of RMOA and select 3 responses on Greedy Diversity Embedding Selection, using the same small model in each layer for consistency. To enhance the diversity and creativity of model outputs, we introduced different role-playing mechanisms \citep{jinxin2023cgmi} for the models. We employed the open-source BGE-m3  \citep{multim3} model for embeddings, and the same model for residual extraction and aggregation. 
Since the MoA and SMoA papers did not conduct experiments on small models (e.g., llama3.1-8B-Instruct), the results presented in this section are derived from our own tests. To ensure the reliability and consistency of the results, we used the same prompts, sampling temperature, and max\_tokens across all datasets. In terms of inference, we employed the vllm \citep{kwon2023efficient}  framework to enhance inference speed, which may result in minor differences compared to existing studies.

\begin{table*}[!htb]
\centering
{\fontsize{10pt}{10pt}\selectfont
\setlength{\tabcolsep}{10pt}
\begin{tabular}{{l l l l l l}}
\toprule[1.5pt]
\textbf{Model} & \textbf{AlpacaEval 2.0} & \textbf{MATH} & \textbf{CRUX} & \textbf{MMLU-r} & \textbf{Average} \\
\midrule
Qwen2.5-7B-Instruct & 37.94 & 74.94 & 57.31 & 69.90 & 60.02\\
\phantom{--}+MoA & 31.77 & 75.28 & 56.81 & 62.70 &  $56.64_{\textcolor{red}{\downarrow \textsubscript{5.63\%}}}$\\
\phantom{--}+SMoA & 40.79 & 76.98 & 59.93 & \textbf{72.00 } &$62.43_{\textcolor{red}{\uparrow \textsubscript{4.02\%}}}$\\
\rowcolor{gray!20}
\phantom{--}+RMoA & \textbf{41.01} & \textbf{77.20} & \textbf{61.00} &  71.80 & $\textbf{62.75}_{\textcolor{red}{\uparrow \textsubscript{4.55\%}}}$\\
\noalign{\vskip 3pt}
\hdashline
\noalign{\vskip 5pt}

Gemma2-9B-Instruct & 45.15 & 36.64 & 47.50 & 63.90 & 48.30\\
\phantom{--}+MoA & 42.73 & 48.92 & \textbf{51.50} & 65.73 & $52.22_{\textcolor{red}{\uparrow \textsubscript{8.12\%}}}$ \\
\phantom{--}+SMoA & 43.23 & 49.96 & 51.25 & 65.80  & $52.56_{\textcolor{red}{\uparrow \textsubscript{8.82\%}}}$\\
\rowcolor{gray!20}
\phantom{--}+RMoA & \textbf{45.61} & \textbf{50.44} & 50.50 &  \textbf{66.10} & $\textbf{53.16}_{\textcolor{red}{\uparrow \textsubscript{10.06\%}}}$ \\
\noalign{\vskip 3pt}
\hdashline
\noalign{\vskip 5pt}

Llama3.1-8B-Instruct & 22.93 & 48.18 & 40.62 & 58.60 & 42.58\\
\phantom{--}+MoA & 30.43 & 50.60 & \textbf{46.12} & 55.10 & $45.56_{\textcolor{red}{\uparrow \textsubscript{7.00\%}}}$ \\
\phantom{--}+SMoA & 31.99 & 51.20 & 44.81 & 60.86  & $47.21_{\textcolor{red}{\uparrow \textsubscript{10.87\%}}}$ \\
\rowcolor{gray!20}
\phantom{--}+RMoA & \textbf{32.86} & \textbf{52.10} & 42.65 &  \textbf{61.63} & $\textbf{47.41}_{\textcolor{red}{\uparrow \textsubscript{11.10\%}}}$ \\
\noalign{\vskip 3pt}
\hdashline
\noalign{\vskip 5pt}

GPT-4o & 55.18 & 76.60 & 75.80 & 83.73 & 72.83\\
\phantom{--}+MoA & 60.55 & 80.08 & 86.66& 85.80 & $78.27_{\textcolor{red}{\uparrow \textsubscript{7.47\%}}}$ \\
\phantom{--}+SMoA & 56.24 & 78.08 & 86.93 & 84.94  & $76.55_{\textcolor{red}{\uparrow \textsubscript{5.11\%}}}$\\
\rowcolor{gray!20}
\phantom{--}+RMoA & \textbf{63.29}  & \textbf{81.16} & \textbf{87.37} &  \textbf{86.67} & $\textbf{79.62}_{\textcolor{red}{\uparrow \textsubscript{9.32\%}}} 
$ \\

\bottomrule[1.5pt]
\end{tabular}
}
\caption{Experimental results of various methods on the AlpacaEval2.0, MATH, CRUX, and MMLU-redux datasets, evaluated using the original benchmark metrics.}
\label{tab:Table 1}
\end{table*}

\subsection{Results}
As shown in Table~\ref{tab:Table 1},we conducted a comprehensive comparison of various MoA methods across multiple datasets, including AlpacaEval2.0, MATH, CRUX, and MMLU-redux.

\textbf{MATH}\quad On the MATH benchmark, our method significantly improves model performance. Specifically, the Qwen2.5-7B-Instruct model achieves a +2.26\% absolute accuracy increase, Gemma2-9B-Instruct shows a breakthrough improvement of +13.8\%, and Llama3.1-8B-Instruct sees a +3.92\% improvement. Notably, even on the larger GPT-4o model, we observe a significant +4.56\% gain, demonstrating the exceptional performance of our method in mathematical reasoning tasks.

\textbf{CRUX}\quad On the CRUX dataset, our method achieves optimal performance with Qwen2.5-7B-Instruct (+3.69\%) and GPT-4o (+11.57\%). While Gemma2-9B-Instruct and Llama3.1-8B-Instruct also show positive gains, their improvements are slightly lower compared to traditional MoA methods ($\Delta=1.2\%-1.8\%$). This suggests that in code understanding tasks, the introduction of redundant tokens can positively influence performance by enhancing context modeling.

\textbf{MMLU-redux}\quad In the MMLU-redux multi-domain knowledge evaluation, the RMoA method results in an average accuracy increase of +2.51\%. The SMoA method also shows a +1.86\% improvement. However, for MoA on smaller models (e.g., Qwen2.5-7B-Instruct and Llama3.1-8B-Instruct), performance significantly drops below the baseline ($\Delta=-3.50\%$~$\sim$~$-7.20\%$). This outcome verifies that redundant information may interfere with reasoning processes in knowledge-intensive tasks.

\textbf{AlpacaEval 2.0}\quad Due to experimental resource constraints, we used the official GPT-4o-mini as the evaluator. The results indicate that our method consistently achieves optimal performance across models of varying scales, particularly with a notable +8.11\% improvement on the GPT-4o model. It is worth noting that although the improvement for Gemma2-9B-Instruct is relatively modest ($\Delta=+0.46\%$), both SMoA and traditional MoA methods experience performance degradation on this model ($\Delta=-1.92\%$~$\sim$~$-2.42\%$).

\section{Analysis}
In this section, we conduct comprehensive experiments to thoroughly investigate the mechanisms of RMoA. 
The experiments are primarily divided into ablation studies, cost analysis, and case studies.

\subsection{Ablation Study}
We conducted ablation experiments by fixing each layer's model to 6 Qwen2.5-7B-Instruct to systematically analyze the contribution of each RMoA component to model performance. 
In the following sections, we will analyze the impact of each component in detail.

\textbf{The number of responses selected through greedy diversity embedding is a critical hyperparameter}.  As shown in Table~\ref{tab:Table 2}, on the MATH and CRUX datasets, model performance increases when the response number \( K \) is 2 or 3, but decreases at \( K \) values of 4 and 5. Similarly, on the MMLU-redux dataset, performance improves at \( K \) values of 2, 3, and 4, but declines at 5. Therefore, selecting \( K = 3 \) strikes a balance between model performance and computational cost. Notably, in SMoA's response selection process, \( K = 3 \) also proves to be an optimal choice.
\begin{table*}[t]
\centering
\begin{minipage}{.42\linewidth}
    \centering
    \small
    \setlength{\tabcolsep}{4pt}
    \begin{tabular}{lcccc}
        \toprule[1.5pt]
         & \textbf{Math} & \textbf{CRUX} & \textbf{MMLU-r} & \textbf{Cost}\\
        \midrule
        MoA  & 75.28 & 57.31 &  62.70 &176.59 \\ 
        \noalign{\vskip 1pt}
        \hdashline
        \noalign{\vskip 3pt}
        RMOA & & &\\
        \phantom{--}w/ K=2 & 76.24 & 58.12 & 71.30 & \textbf{104.47} \\
        \phantom{--}w/ K=3 & \textbf{77.20} & \textbf{61.00} & 71.80 & 121.55\\
        \phantom{--}w/ K=4   & 76.82 & 60.06 & \textbf{72.26} &146.30\\
        \phantom{--}w/ K=5   & 76.78 & 59.87  & 72.16 &178.62\\
        \bottomrule[1.5pt]
    \end{tabular}
    \caption{Hyperparameter analysis of the response count \(K\) in Greedy Diversity Embedding Selection for Qwen2.5-7B-Instruct.}
    \label{tab:Table 2}
\end{minipage}\hfill
\begin{minipage}{.51\linewidth}
    \centering
    \setlength{\tabcolsep}{4pt}
    {\fontsize{9}{10}\selectfont
    \begin{tabular}{lcccc}
    \toprule[1.5pt]
    \textbf{Model} & \textbf{Extractor} & \textbf{Aggreator} \\ 
    \hline 
    \noalign{\vskip 3pt}
    Llama2-7B-Instruct & 49.26 & 49.30 \\ 
    Llama3.1-8B-Instruct & 52.10 & 52.10 \\
    Qwen2.5-72B-Instruct & 53.38 & \textbf{80.16} \\
    DeepSeek-R1-Distill-Llama-70B & \textbf{56.12} & 53.52 \\ 
    \bottomrule[1.5pt]
    \end{tabular}
    }
    \caption{Evaluating the Impact of Models as Residual Extractors and Aggregators on MATH Dataset. LLaMA-3.1-8B-Instruct acts as the aggregator when evaluating extractors, and vice versa. The setup uses four RMoA layers with LLaMA-3.1-8B-Instruct as the proposer.}
    \label{tab:Table 3}
\end{minipage}
\end{table*}

\begin{figure}[ht]
  \includegraphics[width=\linewidth]{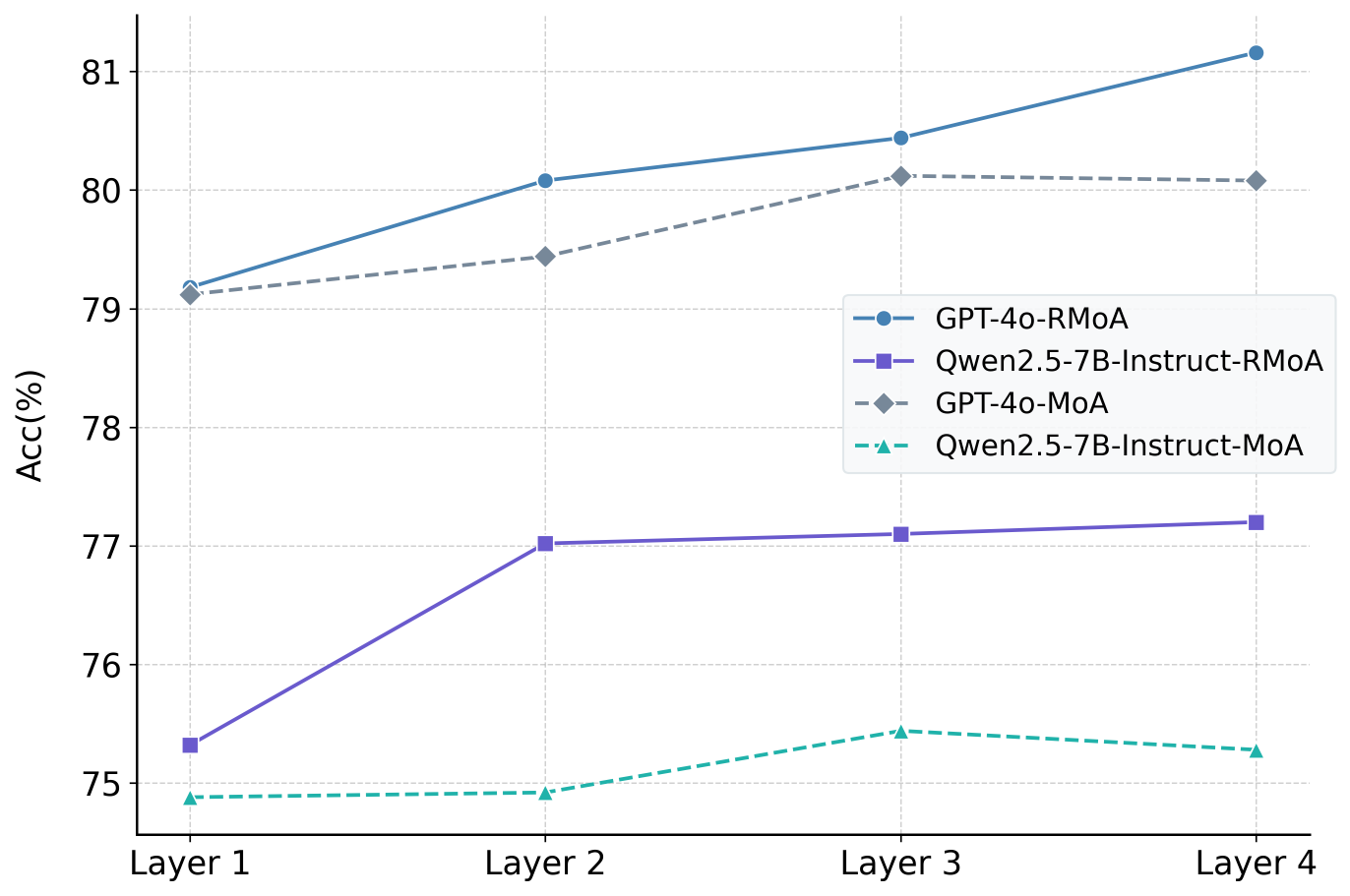}
  \caption{Performance comparison of RMoA and MoA across different layer counts on GPT-4o and Qwen2.5-7B-Instruct.}
  \label{fig:Figure 3}
\end{figure}

\begin{figure}[ht]
  \includegraphics[width=\linewidth]{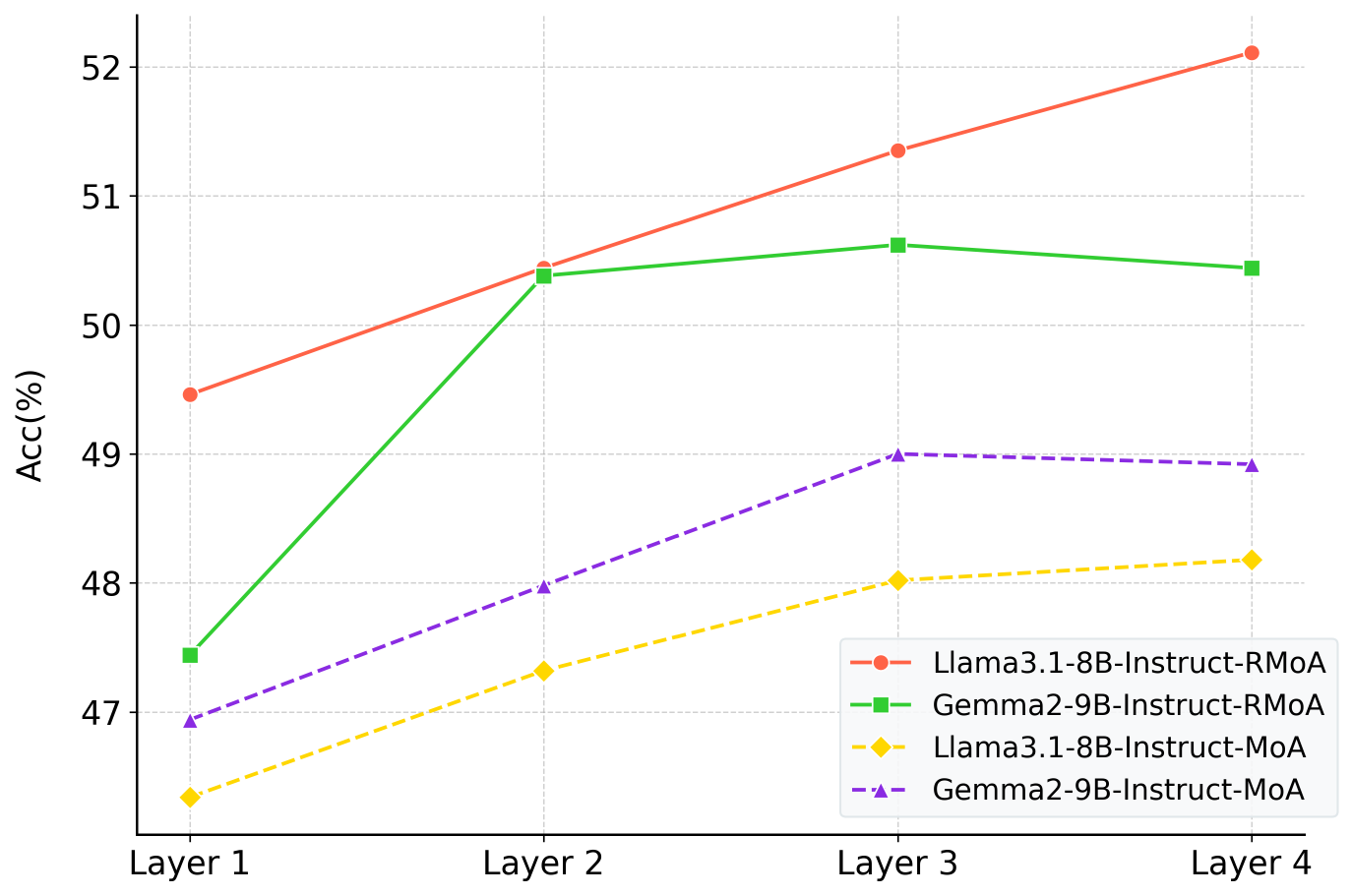}
  \caption{Performance comparison of RMoA and MoA across different layer counts on Llama3.1-8B-Instruct and Gemma2-9B-Instruct.}
  \label{fig:Figure 3-2}
\end{figure}

\begin{table}[!htb]
\centering
\small 
\begin{tabular}{lcccc}
\toprule[1.5pt]
 & \textbf{Math} & \textbf{CRUX} & \textbf{MMLU-r} & \textbf{Cost}\\
\midrule
MoA  & 75.28 & 57.31 &  62.70 & 176.59\\ 
\hdashline 

RMOA & \textbf{77.20} & \textbf{61.00} & 71.80  & 121.55\\
\phantom{--}w/o ES & 76.90 & 60.37 & \textbf{72.10} &  207.23\\
\phantom{--}w/o RA  & 75.90 & 59.37 & 71.60 & 90.37\\
\phantom{--}w/o AT  & 77.10 & 59.62  & 71.70 & 138.56\\
\bottomrule[1.5pt]
\end{tabular}
\caption{Ablation study results with Qwen2.5-7B-Instruct. ES, RA, and AT correspond to Greedy Diversity Embedding Selection, Residual Extraction Agent, and Adaptive Termination.The cost metric refers to the total dollar expenditure of the method across the three datasets. It is calculated based on the Together API's pricing model, which charges \$0.30 per 1 million tokens.}
\label{tab:Table 4}
\end{table}

\begin{figure*}[!htbp]  
  \centering
  \begin{subfigure}{0.48\textwidth}  
    \includegraphics[width=\linewidth]{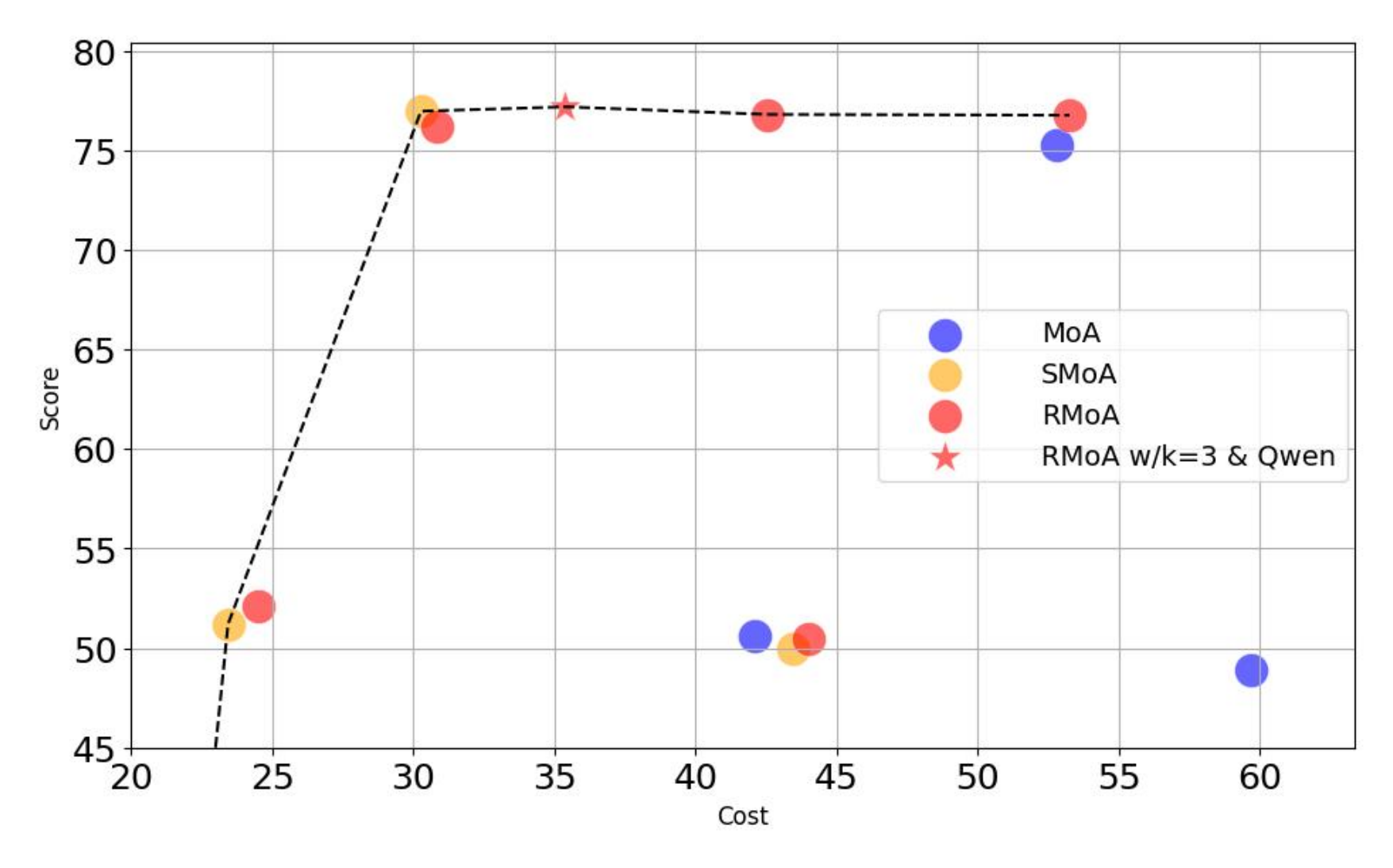}
    \caption{Performance vs Cost}
    \label{fig:Figure 4(a)}
  \end{subfigure}
  \hfill  
  \begin{subfigure}{0.48\textwidth}
    \includegraphics[width=\linewidth]{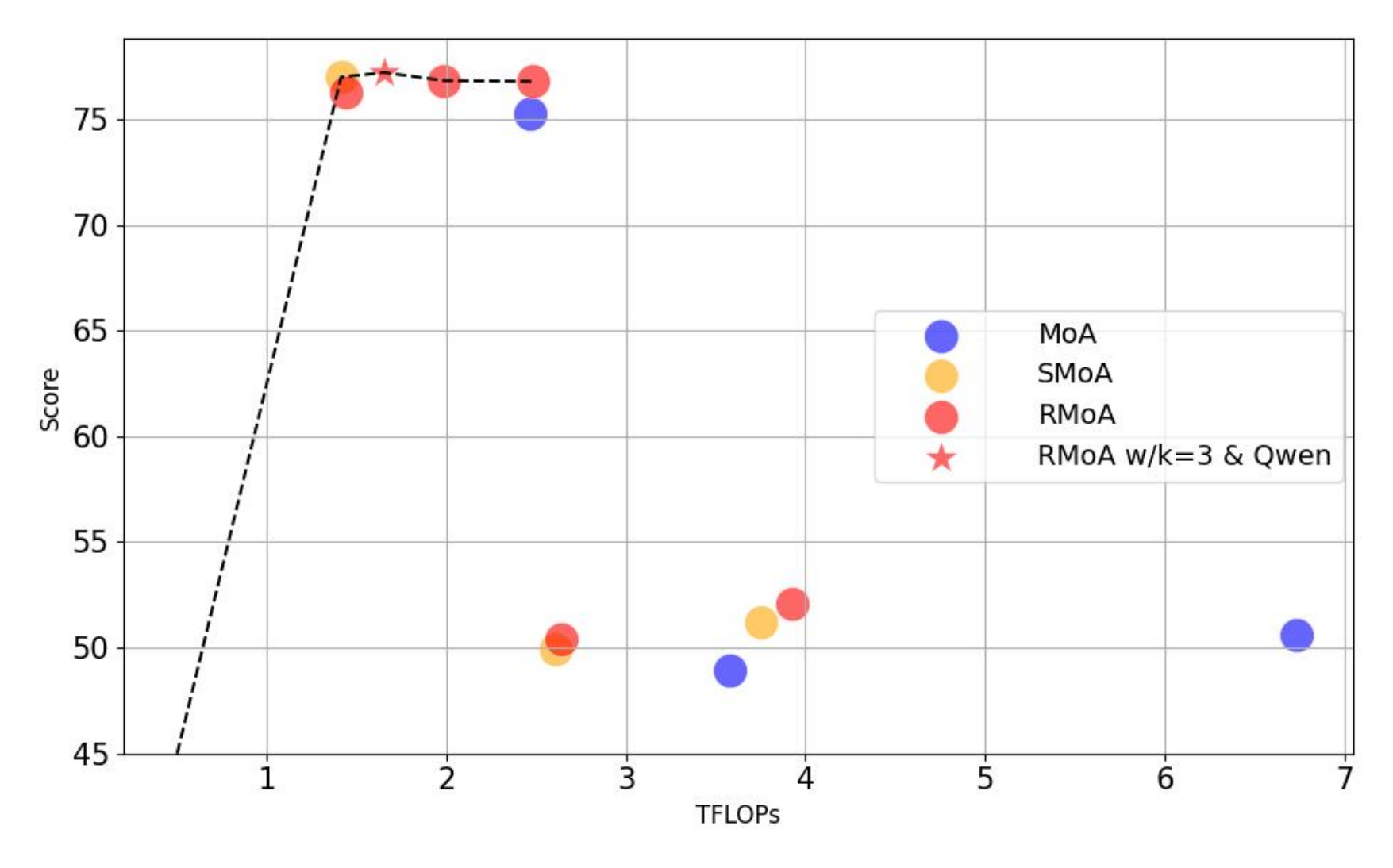}
    \caption{Performance vs TFLOPs}
    \label{fig:Figure 4(b)}
  \end{subfigure}
  \caption{Comparison of Performance Metrics.Figure 5(a) illustrates the token cost incurred by Qwen2.5-7B-Instruct, Llama3.1-8B-Instruct, and Gemma2-9B-Instruct on the MATH dataset using MoA, SMoA, and RMoA. The cost per token is based on pricing from the Together API.Figure 5(b) shows the TFLOPs expenditure for the same models and methods. Due to differences in inference system costs, we estimate TFLOPs based on the model parameters.}
  \label{fig:experiments}
  \vspace{-3pt}
\end{figure*}

\textbf{Models with strong comprehensive capabilities can enhance the effectiveness of residual extraction and aggregation.} \quad
In our investigation of the impact of different model capabilities on residual extraction and aggregation, we selected Llama3.1-8B-Instruct as the base model for our experiments, with all proposers being Llama3.1-8B-Instruct. For residual extraction, as shown in Table~\ref{tab:Table 3}, we fixed the residual aggregator as the Llama3.1-8B-Instruct model and varied the residual extractor model. The results indicate that using the more capable Qwen2.5-72B and Deepseek-R1-Distill-Llama-70B models improved performance on the MATH task by 1.28\% and 4.02\%, respectively. In contrast, the less capable Llama2-7B-Instruct led to a performance decrease of 2.84\%. A similar trend was observed for residual aggregation. Notably, when using Qwen2.5-72B-Instruct for aggregation, performance increased significantly by 28.06\%. This improvement may be attributed to the aggregator not only referencing the information provided by the extractor and residuals but also leveraging its own knowledge for aggregation. This phenomenon aligns with previous findings \citep{Xie2024MindScopeEC,bai2024fairmonitor} on cognitive biases and the curse of knowledge in Large language models.The appendix ~\ref{sec:appendix-model aggregation} contains supplementary details

\textbf{RMoA demonstrates a stronger capability for deep-level iteration. }As shown in Figure~\ref{fig:Figure 3} and Figure~\ref{fig:Figure 3-2}, the performance of different models on the MATH dataset continuously improves with the increase in layers, whereas MoA exhibits varying degrees of decline across all models. This further indicates that MoA may generate hallucinations during the iteration process, leading to originally correct answers becoming incorrect. In contrast, RMoA completes the task more effectively, showcasing its potential for deeper-level iteration compared to MoA.

\textbf{Adaptive Termination mitigates performance degradation caused by hallucinations due to excessive iterations and helps reduce costs.} As shown in Table~\ref{tab:Table 4},  Adaptive Termination led to varying degrees of improvement across different datasets, with the most notable increase of 1.38\% observed on the CRUX dataset. This improvement is likely because smaller models may generate hallucinations when they continue to update responses after already providing correct answers. Additionally, the implementation of adaptive early stopping resulted in a cost savings of \$17.01.
\subsection{Budget Analysis}
\textbf{Cost Efficiency Analysis} \quad
In Figure~\ref{fig:Figure 4(a)}, we present the relationship between ACC and the total inference cost on the MATH benchmark. Since we are using local inference, accurately quantifying specific costs is challenging. Therefore, we utilize model pricing from the API website for our calculations. The graph illustrates a Pareto frontier, indicating that certain models achieve a better balance between cost and performance. Models closer to the Pareto frontier are more cost-effective. Specifically, our RMoA, by selecting three differentiated responses and employing Qwen2.5-7B-Instruct as the model for all agents, achieves the optimal configuration. Compared to MoA with the same model configuration, RMoA improves performance by 1. 92\% while only 68. 83\% of MoA.

\textbf{TFLOPs Analysis }\quad
Due to the varying latencies caused by different inference systems, we use the number of TFLOPs as a proxy for latency.In Figure~\ref{fig:Figure 4(b)},the chart describes the relationship between ACC and the number of TFLOPs, where a Pareto frontier is also observable. Models on this frontier effectively utilize their computational resources to maximize accuracy on MATH. Specifically, compared to MoA with the same configuration, RMoA achieves a 1.92\% increase in accuracy while reducing TFLOPs by nearly 31.88\%.

\subsection{Case Study}

By demonstrating the effectiveness of greedy diversity selection and residual extraction (more details in Figures ~\ref{fig: case1}, ~\ref{fig: case2}, and ~\ref{fig: case3}), we observe that the responses from the four models contain a large amount of homogeneous content. After applying greedy diversity selection, GPT-4o and Qwen2.5-7B-Instruct were chosen. These two responses encompass the vast majority of the content from all responses, highlighting the effectiveness of the greedy diversity selection method. Furthermore, by performing residual extraction of the responses selected from two consecutive rounds, we identified additional information and detail discrepancies related to the questions. This provides a solid foundation for subsequent residual aggregation.

\section{Conclusion}
This paper introduces the RMoA Framewrok, which ultilizes iterative collaboration to improve MAS capabilities. We propose Greedy Differential Embedding Selection, Residual Agent, and Adaptive Termination Mechanism to achieve diversity maximization and residual compensation. The proposed RMoA alleviates the problems of high computational overhead, information loss and robustness of traditaional MoA architecture. We conduct extensive evaluations across a variety of tasks and explore the potential of RMoA through ablation studies and cost analysis.

\section*{Limitation}
In this work, we introduce residuals to mitigate information loss between layers, enabling our method to achieve performance gains even at deep layers. However, due to time and cost constraints, we have not yet explored the performance limits of our approach. In future work, we aim to evaluate the performance limits of various models across different depths and analyze the scaling laws that govern these limits.

\section*{Acknowledgements}
This work is supported by the National Natural Science Foundation of China (Grant No. 62207013).
\bibliography{custom}

\clearpage
\appendix

\section{Prompt Design}
In this section, we provide all the prompts used by RMoA in the experiments. Specifically, Figure~\ref{fig: prompt1} illustrates the prompts used by RMoA's residual extraction and residual aggregation agents. Meanwhile, Figure~\ref{fig: prompt2} shows the prompts used in the MoA and SMoA baseline models, which were sourced from their official GitHub projects.

Furthermore, to enhance the distinctiveness of model output, we assign different role-playing descriptions \citep{jinxin2023cgmi} to each dataset. Specifically, Figure~\ref{fig: prompt2} presents the role prompts for AlpacaEval 2.0, Figure ~\ref{fig: prompt3} presents the role prompts for CRUX, and Figures~\ref{fig: prompt4} and ~\ref{fig: prompt5} display the role prompts for MATH and MMLU-Redux, respectively.

For different datasets, we employ different reasoning modes to optimize performance. As shown in Figures ~\ref{fig: prompt6} and ~\ref{fig: prompt7}, we used a few-shot approach for the MATH data set. Figure ~\ref{fig: prompt7} illustrates the use of Chain-of-Thought (CoT) reasoning for the CRUX dataset. Due to the extensive nature of the CoT content for MMLU-Redux, the prompts can be found in the code's prompt file. Lastly, for AlpacaEval 2.0, we adopted a zero-shot approach.

\begin{table}[!htbp]
\centering
\begin{tabular}{lc}
\toprule[1.5pt]
\textbf{Benchmark} & \textbf{Method} \\ \hline 
MATH & Few-Shot \\ 
CRUX & CoT \\
MMLU-redux & CoT \\
AlpacaEval 2.0& Zero-Shot \\ 
\bottomrule[1.5pt]
\end{tabular}
\caption{Inference Modes for Different Datasets}
\label{tab:Table 5}
\end{table}

\section{Acknowledgment of AI Assistance in Writing and Revision}
We utilized GPT-4o for paper refinement and grammar correction.

\section{More results}
\subsection{Comparison of Residual Quantization Methods}
\label{sec:appendix-model Comparison of Residual}
\begin{itemize}
    \item \textbf{LLM-based judgment:} LLM judges whether the key information has changed significantly between any previous-round (\( R_{t-1}^i \)) and any current-round (\( R_t^j \)) reference responses:
        \[
        R_t = \text{Cat}(R_{t,i=1}^{iK}, R_{t-1,j=1}^{jK}),
        \]
        \[
        \Delta R = \text{ResidualExtractor}(R_t, \text{prompt}).
        \]
          If the residual extractor determines that no residual information exists, the iteration terminates; otherwise, it continues.
        
        \item \textbf{Similarity threshold:} All pairwise cosine similarities between any previous-round (\( R_{t-1}^i \)) and any current-round (\( R_t^j \)) reference responses exceed \( \theta \):
        \[ \text{Sim}(R_{t-1}^i, R_t^j) = \frac{\text{Emb}(R_{t-1}^i)^\top \text{Emb}(R_t^j)}{\|\text{Emb}(R_{t-1}^i)\|_2 \|\text{Emb}(R_t^j)\|_2}, \]
        \[
        \forall i, j, \quad \text{Sim}(R_{t-1}^i, R_t^j) > \theta.
        \]

        \item \textbf{Variance Metric:} \\
The total variance of pairwise similarities is bounded by \( \sigma^2 \):
\[
\sum_{i,j} \left( \text{Sim}(R_{t-1}^i, R_t^j) - \mu \right)^2 < \sigma^2,
\]
where \( \text{Sim}(\cdot, \cdot) \) is the cosine similarity, and \( \mu \) is the mean similarity:
\[
\mu = \frac{1}{K^2} \sum_{i,j} \text{Sim}(R_{t-1}^i, R_t^j).
\]

\end{itemize}
Table ~\ref{tab:convergence_criteria} shows LLM-based judgments outperform quantitative methods. Analysis reveals semantic similarity and residual convergence are weakly correlated in reasoning tasks—for instance, the embedding representations derived from collaborative dialogue may fail to capture critical textual information, which fundamentally limits the improvements achievable through basic semantic similarity measures. Thus, quantitative metrics often miss critical reasoning differences captured robustly by LLMs.
\begin{table}[!ht]
    \small
    \centering
    \begin{tabular}{l l l}
        \toprule
        \textbf{Criterion Type} & \textbf{Threshold} & \textbf{Accuracy} \\
        \midrule
        LLM judgment & -- & \textbf{80.20\%} \\
        similarity thresholds & $\theta = 0.7$ & 76.20\% \\
        & $\theta = 0.8$ & 76.60\% \\
        & $\theta = 0.9$ & 78.80\% \\
        Variance metric & $\sigma^2 = 1 \times 10^{-3}$ & 77.40\% \\
        & $\sigma^2 = 1 \times 10^{-3}$ & 79.40\% \\
        & $\sigma^2 = 1 \times 10^{-3}$ & 78.40\% \\
        \bottomrule
    \end{tabular}
    \caption{Performance comparison of different convergence criteria on Math500 dataset}
    \label{tab:convergence_criteria}
\end{table}
\subsection{The impact of different embedding models on RMOA}
\label{subsec:applendix embedding models}
To better understand the impact of different embedding models on RMoA, we tested three embedding models—BGE-m3\citep{multim3}, SGPT-1.3B-mean-nli\citep{muennighoff2022sgpt}, and multilingual-e5-large\citep{wang2024multilingual}—using qwen2.5-7b-Instruct on MATH500\citep{lightman2023lets}. Results (Table ~\ref{tab:embedding_accuracy}) show minimal performance variation (0.6\%), demonstrating the robustness of our method to the choice of embedding model.

\begin{table}
\centering
\begin{tabular}{lc}
\toprule
\textbf{Embedding Model} & \textbf{Accuracy} \\
\midrule
BGE-m3-dense & \textbf{0.806} \\
SGPT-1.3B-mean-nli & 0.796 \\
multilingual-e5-large & 0.790 \\
\bottomrule
\end{tabular}
\caption{Accuracy comparison of embedding models}
\label{tab:embedding_accuracy}
\end{table}
\subsection{Performance of RMoA on different models with different numbers of layers.}
\begin{table}[!htbp]

\centering
\small
\setlength{\tabcolsep}{4pt}
\begin{tabular}{lccc}
    \toprule[1.5pt]
     & \textbf{Math} & \textbf{CRUX} & \textbf{MMLU-r} \\
    \midrule
    RMOA & & &\\
    \phantom{--}Layer 1 & 75.32 & 57.12 & 69.96  \\
    \phantom{--}Layer 2 & 77.02 & 58.50 & 70.83 \\
    \phantom{--}Layer 3   & 77.10 & 59.50 & 70.9 \\
    \phantom{--}Layer 4  & 77.14 & 60.01  & 70.93 \\
    \phantom{--}Layer 5   &\textbf{ 77.26} & 60.05 & 71.80 \\
    \phantom{--}Layer 6  & 77.20 & \textbf{61.00 }& \textbf{71.80} \\
    \bottomrule[1.5pt]
\end{tabular}
\caption{The benchmark performance of Qwen2.5-7B-Instruct at different layers on RMoA.}
\label{tab:Table ap 1}
\end{table}

\begin{table}[!htbp]

\centering
\small
\setlength{\tabcolsep}{4pt}
\begin{tabular}{lccc}
    \toprule[1.5pt]
     & \textbf{Math} & \textbf{CRUX} & \textbf{MMLU-r} \\
    \midrule
    RMOA & & &\\
    \phantom{--}Layer 1 & 79.18 & 86.68 & 83.73  \\
    \phantom{--}Layer 2 & 80.08 & 86.81 & 86.06 \\
    \phantom{--}Layer 3   & 80.44 & 87.18 & 86.56 \\
    \phantom{--}Layer 4  & 81.16 & 87.17  & 86.62 \\
    \phantom{--}Layer 5   & 81.32 & 87.37 & \textbf{86.68} \\
    \phantom{--}Layer 6  &\textbf{ 81.34} & \textbf{87.37}  & 86.67 \\
    \bottomrule[1.5pt]
\end{tabular}
\caption{The benchmark performance of GPT-4o at different layers on RMoA.}
\label{tab:Table ap 2}
\end{table}

As shown in ~\ref{tab:Table ap 1} and ~\ref{tab:Table ap 2},the benchmark performance of Qwen2.5-7B-Instruct and GPT-4o across different layers on the RMoA framework, as shown in the tables, indicates a trend of increasing performance with deeper layers. For Qwen2.5-7B-Instruct, there is a consistent improvement across the MATH, CRUX, and MMLU-r benchmarks, with CRUX showing a notable increase from 57.12\% at Layer 1 to 61.00\% at Layer 6. Similarly, GPT-4o demonstrates an upward trend, with MATH scores rising from 79.18\% at Layer 1 to 81.34\% at Layer 6, and comparable improvements in CRUX and MMLU-r benchmarks. These results suggest that the RMoA framework effectively mitigates information loss between layers, leading to enhanced performance in deeper layers, and highlighting the potential for further exploration into layer depth optimization to achieve even greater performance gains.

\subsection{Quantitative evaluation of hallucination rate}
In our work, hallucination rate is defined as: among entries that received correct responses in the previous round, the proportion that produce incorrect responses in the current round. The calculation method is: number of previously correct but now incorrect responses divided by total number of previously correct responses.

As shown in Table ~\ref{tab:hallucination_rate}, our adaptive termination (AT) mechanism demonstrates significant effectiveness on the MATH dataset with both Qwen2.5-7B-Instruct and GPT-4o models, reducing hallucinations by average margins of 3.77\% and 2.95\% per round respectively compared to non-AT approaches. This improvement occurs because AT immediately halts computation upon correct answer generation, thereby eliminating unnecessary reasoning steps that might otherwise degrade output quality through excessive modifications or introduced confusion.

\begin{table*}
\centering
\begin{tabular}{lrrrrr}
\toprule
Model & Round 2 & Round 3 & Round 4 & Round 5 & Round 6 \\
\midrule
Qwen2.5-7B-Instruct w/o AT & 5.91 & 5.17 & 5.28 & 5.73 & 4.85 \\
Qwen2.5-7B-Instruct & \textbf{2.13} & \textbf{1.04} & \textbf{2.01} & \textbf{1.91} & \textbf{1.00} \\
GPT-4o w/o AT & 4.33 & 3.98 & 4.11 & 4.77 & 5.22 \\
GPT-4o & \textbf{2.55} & \textbf{1.29} & \textbf{1.79} & \textbf{1.02} & \textbf{0.98} \\
\bottomrule
\end{tabular}
\caption{Hallucination rate comparison (in \%) with and without AT across conversation rounds.}
\label{tab:hallucination_rate}
\end{table*}

\subsection{Model Aggregation Case Study}
\label{sec:appendix-model aggregation}
Qwen-2.5-72B-Instruct demonstrates superior performance as a residual aggregator (80.16 vs 53.52) because it combines reference information with its own knowledge during aggregation. In contrast, DeepSeek-R1-Distill-Llama-70B solely relies on reference information generated by smaller models.

As shown in Figure ~\ref{fig:model_aggregation_case}, we used llama3.1-8b-instruct to generate reference responses and extract residuals, then compared aggregation performance between Qwen2.5-72b-instruct and DeepSeek-R1-Distill-Llama-70B.As evidenced by the comparative results of the two models, Qwen2.5-72B-Instruct demonstrates comprehensive autonomous reasoning when aggregating reference responses, whereas DeepSeek-R1-Distill-Llama-70B exhibits explicit filtering behavior during response aggregation.

\subsection{Method Effectiveness Validation on Larger-Scale Parameter Models}
We have conducted experiments on more affordable and powerful models. We organized additional experiments with Qwen2.5-72B-Instruct , Deepseek-v3\citep{liu2024deepseek} and Deepseek-R1\citep{guo2025deepseek}.Deepseek-R1's long chain-of-thought reasoning process results in slower inference speed. Due to time constraints, we randomly selected 41 of the most challenging Level5 problems from MATH500 for testing. Meanwhile, we conducted full dataset testing for both DeepSeek-V3 and Qwen2.5-72B-Instruct models.

Table~\ref{tab:math500_performance_vertical} demonstrates that the RMoA framework delivers substantial accuracy gains, particularly in high-complexity tasks. On the full MATH500 dataset, RMoA boosts Qwen2.5-72B’s performance by 7.8\% (80.00\% → 87.80\%) and Deepseek-v3 by 4.2\% (88.20\% → 92.40\%). Even on the Level5 subset (Table 5), improvements persist: Qwen2.5-72B gains 7.32\% (60.97\% → 68.29\%), while Deepseek-v3 rises 4.87\% (73.17\% → 78.04\%). These results confirm that RMoA enhances model capabilities beyond baseline performance, especially in rigorous mathematical reasoning, where incremental gains are notoriously difficult to achieve.

Even as cheaper, more powerful models emerge (e.g., DeepSeek-R1), the framework can amplify their potential. For instance, applying RMoA to DeepSeek-R1 elevates its Level5 accuracy from 78.04\% to 82.92\% (Table ~\ref{tab:math500_level5_performance}), \textbf{proving that the framework complements — rather than competes with — advances in base models.} By dynamically optimizing computational resources (e.g., reducing redundant steps) and maintaining a lightweight architecture, RMoA ensures sustained cost-effectiveness. Its adaptability to new models and tasks, as evidenced by cross-table consistency, positions it as a scalable solution where efficiency and performance gains compound over time.

\begin{table}
\centering
\begin{tabular}{lc}
\hline
\textbf{Model} & \textbf{Accuracy} \\
\hline
Qwen2.5-72B (4-shot) & 80.00\% \\
Qwen2.5-72B-MoA & 84.20\% \\
Qwen2.5-72B-RMoA & \textbf{87.80\%} \\
Deepseek-v3 (4-shot) & 88.20\% \\
Deepseek-v3-MoA & 89.40\% \\
Deepseek-v3-RMoA & \textbf{92.40\%} \\
\hline
\end{tabular}
\caption{Performance comparison on MATH500 (Vertical Layout). The framework configuration adopts 6 layers with 6 proposers, greedy diversity K=3, temperature set to 0.7, and max\_token set to 1024.}
\label{tab:math500_performance_vertical}
\end{table}

\begin{table}
\centering
\begin{tabular}{lcc}
\hline
\textbf{Method} & \textbf{4-shot} & \textbf{RMoA} \\
\hline
DeepSeek-R1 & 78.04\% & \textbf{82.92\%} \\
Qwen2.5-72B & 60.97\% & 68.29\% \\
GPT-4o & 34.14\% & 41.46\% \\
Deepseek-v3 & 73.17\% & 78.04\% \\
\hline
\end{tabular}
\caption{Performance comparison on MATH500 Level5 Subset - 41 Samples.}
\label{tab:math500_level5_performance}
\end{table}
\begin{table}
\centering
\begin{tabular}{lcc}
\hline
\textbf{Metric} & \textbf{Accuracy} & \textbf{Avg Turn} \\
\hline
Turn1 & 97.2\% & 1\\
Turn2 & 98.6\% & 2\\
Turn3 & 98.6\% & \textbf{2.01}\\
Turn4 & 98.6\% & \textbf{2.01}\\
\hline
\end{tabular}
\caption{Analysis of last-letter concatenation task (vertical format).}
\label{tab:last_letter_vertical}
\end{table}

\begin{table*}[!htbp]
\centering
\begin{tabular}{llcccccccc}
\toprule
Method & Metric & \multicolumn{8}{c}{Turn} \\
\cmidrule(lr){3-10}
& & 1 & 2 & 3 & 4 & 5 & 6 & 7 & 8 \\
\midrule
\multirow{3}{*}{MoA} & Accuracy & 75.60\% & 76.60\% & 78.26\% & 78.53\% & 78.53\% & 78.30\% & 79.10\% & 79.06\% \\
& Cost & 31.41 & 67.11 & 102.72 & 138.90 & 175.77 & 213.44 & 251.73 & 290.59 \\
& $\Delta$Acc & 1.0\% & 1.2\% & 1.6\% & 2.0\% & 2.4\% & 1.2\% & 1.2\% & 2.4\% \\
\addlinespace
\multirow{3}{*}{RMoA} & Accuracy & 75.80\% & 78.80\% & 79.10\% & 79.20\% & 80.00\% & 80.40\% & 81.16\% & 82.20\% \\
& Cost & 29.82 & 61.18 & 77.81 & 91.32 & 103.36 & 114.60 & 125.34 & 135.65 \\
& $\Delta$Acc & 1.0\% & 0.4\% & 0.4\% & 0.2\% & 0.4\% & 0.6\% & 0.4\% & 0.4\% \\
\bottomrule
\end{tabular}
\caption{Performance comparison between MoA and RMoA on MATH500. Acc represents the average accuracy of three tests, Cost is the average cost of testing MATH500 across three assessments, and $\Delta$Acc represents the difference between the maximum and minimum values of accuracy in the three tests.}
\label{tab:level comparison}
\end{table*}

\begin{table*}[!htbp]
\centering
\begin{tabular}{llccccc}
\toprule
Method & Metric & \multicolumn{5}{c}{ \( N \) } \\
\cmidrule(lr){3-7}
& & 4 & 5 & 6 & 7 & 8 \\
\midrule
\multirow{2}{*}{MoA} & Accuracy & 77.40\% & 77.85\% & 78.90\% & 79.20\% & 79.60\% \\
& Cost & 82.87 & 110.89 & 138.90 & 173.78 & 208.66 \\
\addlinespace
\multirow{2}{*}{RMoA} & Accuracy & 77.80\% & 78.65\% & 79.20\% & 79.60\% & 80.00\% \\
& Cost & 68.11 & 79.72 & 91.32 & 119.03 & 146.74 \\
\bottomrule
\end{tabular}
\caption{Performance comparison of MoA and RMoA with different processor counts \( N \).}
\label{tab:processor counts}
\end{table*}
\subsection{Mitigating Over-Optimization via Adaptive Termination}

To evaluate the potential risk of over-optimization in residual extraction and aggregation, we designed a controlled experiment using the simple last-letter concatenation task. Leveraging GPT-4o as the proposer, residual extractor, and aggregator within a 4-layer/6-proposer configuration, we systematically tested the adaptive early stopping mechanism enabled by residual information. The experiment compared the proposed RMoA approach against standard MoA, measuring both efficiency (via layer convergence depth) and performance (accuracy). In simpler task scenarios, RMoA’s early stopping mechanism consistently terminated iterations at the second layer (as shown in Table ~\ref{tab:last_letter_vertical}), avoiding unnecessary deeper-layer computations. This shallow convergence not only improved efficiency relative to standard MoA but also maintained performance parity with the original method. Across nearly all test cases, RMoA demonstrated high accuracy while proactively terminating computations, validating that the residual-driven adaptive stopping prevents over-optimization. By quantifying convergence depth and performance metrics, the experiment empirically confirms that residual extraction and aggregation enhance efficiency without compromising accuracy, effectively mitigating over-optimization risks through data-driven termination logic.

\subsection{The impact of framework depth and processor count on performance}

Regarding the impact of framework depth, we tested MoA and RMoA configurations using six processors, a framework depth of eight layers, a temperature setting of 0.7, max tokens of 1024, and a greedy diversity \( K \) of 3. The data in Table ~\ref{tab:level comparison} shows that RMoA surpasses MoA in terms of accuracy, resource efficiency, and stability. In terms of accuracy (Acc), RMoA improves consistently from Turn 1 (75.80\%) to Turn 8 (82.20\%), significantly surpassing MoA's range of 75.60\%-79.06\%, with the gap widening further in deep iterations (e.g., a 3.14\% difference in Turn 8), verifying the residual extraction mechanism's ability to compensate for information loss between layers. In terms of cost, RMoA's cumulative cost at Turn 8 was 135.65, only 46.7\% of MoA's 290.59, with the single-turn cost increase slowing (an average of 15.12 per turn, compared to MoA's 36.88), indicating the dynamic adjustment mechanism effectively suppresses redundant computation in deep layers. Regarding stability (\(\Delta\)Acc), RMoA's fluctuation range remains 0.50\% (e.g., 0.2\% in Turn 4), whereas MoA's fluctuation peaks at 2.4\%, highlighting adaptive termination's strong control over "over-optimization" risks. In summary, RMoA achieves "high precision-low consumption-strong stability" in deep iterative optimization through residual information integration and efficiency regulation, with the data difference further reinforcing its technical advantages.
    
Concerning the impact of the number of processors per layer, we tested MoA and RMoA configurations with processor numbers from 4 to 8, a framework depth of four layers, a temperature of 0.7, max tokens of 1024, and a greedy diversity \( K \) of 3. As Table ~\ref{tab:processor counts} results show, RMoA outperforms MoA in both accuracy (Acc) and resource efficiency (Cost). With the increase in the number of processors, RMoA's accuracy continually improves (e.g., reaching 80.00\% with eight processors, 0.4\% higher than MoA), and the cost reduces significantly (146.74 vs. 208.66). The residual information fusion mechanism effectively alleviates deep information loss, and the dynamic adjustment strategy suppresses redundant computation, with its single-turn cost increase slowing by 58.9\% compared to MoA (15.12 vs. 36.88).

\clearpage
\label{sec:appendix}
\begin{figure*}[!ht]
    \begin{tcolorbox}[
        title={Detailed Instruction},
        width=\textwidth,
        colback=white,
        colframe=black,
        fonttitle=\bfseries,
        fontupper=\normalsize,
        left=10pt, right=10pt, top=12pt, bottom=10pt,
        parbox=false
    ]
        \textbf{Residual Extraction Prompt for RMoA:} 
        
        You are tasked with performing Residuals Simulation in Residual Networks.
        
        \textbf{Tasks:} \\
        Compare the results of multiple model responses from the previous round with those from the current round. Identify specific differences such as content hallucinations, detail discrepancies, or additional information. If no significant differences are found, indicate accordingly. Ensure that only genuine residuals are reported.
        
        \textbf{Chain-of-thought:} \\
        1. Comparison Basis: Perform a one-to-one comparison between each model's response from the previous round and its corresponding response in the current round. \\
        2. Types of Residuals to Identify: \\
           Content Errors (Hallucinations): Factual inaccuracies or fabricated information introduced in the current response. \\
           Detail Discrepancies: Missing details, additional specifics, or changes in the level of detail. \\
           Additional Information: New information or perspectives not present in the previous response. \\
        3. Output Format: \\
           Overall Indicator: Start with "Residuals Detected: Yes" if at least one model has residuals. Use "Residuals Detected: No" if no significant differences are found across all models. \\
           Residual Details: For each model with residuals, provide a concise description of the specific differences. List each model's residual on a separate line, prefixed by the model number for clarity. \\
        4. Authenticity Assurance: Only report actual differences. Do not infer or generate residuals that do not exist. Verify each identified residual to ensure its validity and relevance.
        
        \textbf{Residual Aggregation Prompt for RMoA:}\\ You are the "Residual Aggregator". You have two key inputs: previous response and current-layer residuals. Deliver a well-rounded, error-free, and unbiased final answer that demonstrates thorough integration of all relevant information.
        
        \textbf{Tasks:} \\
        1. Synthesize all residuals into a single, concise, and accurate answer. \\
        2. Integrate Residuals to fill gaps, include alternative views, and correct errors. \\
        3. Evaluate Critically for bias or inaccuracy, ensuring reliability and objectivity. \\
        4. Present Structurally, maintaining clear organization and logical flow.
        
        \textbf{Chain-of-thought:} \\
        1. Review all responses for common points and discrepancies. \\
        2. Draft a unified answer that captures essential information. \\
        3. Incorporate Residuals by adding unique insights or corrections. \\
        4. Finalize the response for clarity, coherence, and impartiality.
    \end{tcolorbox}
    \caption{Agent Prompt Design-Part One}
    \label{fig: prompt1}
\end{figure*}

\label{sec:appendix}
\begin{figure*}[!ht]
    \begin{tcolorbox}[
        title={Detailed Prompt},
        width=\textwidth,
        colback=white,
        colframe=black,
        fontupper=\normalsize,
        fonttitle=\bfseries,
        left=10pt, right=10pt, top=12pt, bottom=10pt,
        parbox=false
    ]
        \textbf{Aggregation for SMoA and MoA:} \\
        You have been provided with a set of responses from various open-source models to the latest user query. Your task is to synthesize these responses into a single, high-quality response. It is crucial to critically evaluate the information provided in these responses, recognizing that some of it may be biased or incorrect. Your response should not simply replicate the given answers but should offer a refined, accurate, and comprehensive reply to the instruction. Ensure your response is well-structured, coherent, and adheres to the highest standards of accuracy and reliability.
        
        \vspace{5pt}
        
        \textbf{Moderator \& Judge Prompt for SMoA:} \\
        You are a moderator. You will be provided with a set of responses from various open-source models to the latest user query. Your task is to carefully and meticulously select [Response Number] responses from them, according to correctness, fluency, relevance, and quality. It is crucial to critically evaluate the information provided in these responses, recognizing that some of them may be biased or incorrect. Additionally, you need to decide whether to end the debate by measuring the consistency between responses and giving an indicator controlling ending the debate or not. The output should be a markdown code snippet formatted in the following schema: "reasoning": str // Logical reasoning behind the chosen response "chosen responses": list // the best [Response Number] response. For example [0, 1] "debate": bool // whether end the debate Question:
        
        \vspace{5pt}
        
        \textbf{AlpacaEval2.0 Role Description Prompt:} \\
        \textbf{Role1:} You are a natural language processing researcher specializing in evaluation methodologies for large language models. You are analytical, detail-oriented, and deeply invested in addressing biases in automated assessment systems. You collaborate with AI ethics teams at tech conferences, publishing papers on robust evaluation metrics and developing protocols to ensure fair model comparisons.
        
        \textbf{Role2:} You are a data scientist with expertise in statistical bias correction and metric design. You are pragmatic, numerically adept, and excel at identifying confounding variables in evaluation frameworks. You work at a machine learning startup, developing novel techniques to isolate model capabilities from superficial factors like response length through rigorous statistical analysis.
        
        \textbf{Role3:} You are an AI alignment engineer focused on instruction following robustness. You are systematic, solution-driven, and passionate about bridging the gap between human intent and model behavior. You participate in developer forums and open-source communities, creating benchmarking tools that test models' ability to handle nuanced real-world instructions.
        
        \textbf{Role4:} You are an educational technology specialist designing AI literacy curricula. You are articulate, pedagogically skilled, and passionate about translating complex evaluation concepts into accessible formats. You collaborate with university AI labs and K-12 educators to create instructional materials that explain model benchmarking principles to diverse audiences.
        
        \textbf{Role5:} You are a cognitive science researcher studying human-AI interaction patterns. You are curious, interdisciplinary, and investigate how people perceive model response quality. You conduct user studies at human-computer interaction conferences, providing empirical insights about the relationship between quantitative metrics and subjective response quality assessments.
    \end{tcolorbox}
    \caption{Agent Prompt Design - Part two}
    \label{fig: prompt2}
\end{figure*}

\label{sec:appendix}
\begin{figure*}[!ht]
    \begin{tcolorbox}[
        title={Detailed Prompt},
        width=\textwidth,
        colback=white,
        colframe=black,
        fonttitle=\bfseries,
        fontupper=\normalsize,
        left=10pt, right=10pt, top=12pt, bottom=10pt,
        parbox=false
    ]
        \textbf{Role6:} You are an open-source model evaluation platform maintainer. You are community-oriented, technically proficient, and dedicated to building transparent benchmarking infrastructure. You coordinate with volunteer developers and researchers to implement standardized testing protocols while maintaining compatibility with diverse model architectures.
        
        \vspace{5pt}
        
        \textbf{CRUX Role Description Prompt:} \\
        \textbf{Role1:} You are a software engineer specializing in code optimization and refactoring. You are analytical, detail-oriented, and passionate about improving code efficiency and readability. You work closely with other developers and software architects, focusing on enhancing existing codebases and ensuring high performance and maintainability. Your role often involves reviewing code, identifying bottlenecks, and implementing solutions to optimize functionality.
        
        \textbf{Role2:} You are a data scientist with expertise in machine learning and predictive modeling. You are innovative, data-driven, and skilled at extracting insights from complex datasets. You collaborate with cross-functional teams, including engineers and product managers, to develop models that predict outcomes and inform decision-making. Your work involves using statistical methods and programming to analyze data and build robust models.
        
        \textbf{Role3:} You are a computer science professor with a focus on programming languages and software development. You are methodical, knowledgeable, and dedicated to teaching and mentoring students. In academic settings, you engage in research and curriculum development, aiming to advance understanding in programming theory and practice. You often participate in conferences and publish papers on software engineering topics.
        
        \textbf{Role4:} You are a technical writer specializing in software documentation. You are precise, communicative, and skilled at translating complex technical concepts into clear and concise documentation. You work closely with developers and product teams to create user manuals, API documentation, and tutorials that help users understand and utilize software effectively. Your role involves ensuring that documentation is accurate, comprehensive, and accessible.
        
        \textbf{Role5:} You are a software developer with a passion for coding and problem-solving. You are creative, resourceful, and enjoy building applications from scratch. You frequently collaborate with designers and product managers to create user-friendly software solutions. Your role involves writing clean, efficient code and staying up-to-date with the latest programming languages and technologies to continuously improve your skills.
        
        \textbf{Role6:} You are a cybersecurity analyst with expertise in software security and vulnerability assessment. You are vigilant, analytical, and dedicated to protecting systems from cyber threats. You work with IT teams to identify and mitigate security risks, conduct penetration testing, and develop strategies to enhance security protocols. Your role involves staying informed about the latest security trends and technologies to safeguard digital assets.
    \end{tcolorbox}
    \caption{Agent Prompt Design - Part Three}
    \label{fig: prompt3}
\end{figure*}

\label{sec:appendix}
\begin{figure*}[!ht]
    \begin{tcolorbox}[
        title={Detailed Prompt},
        width=\textwidth,
        colback=white,
        colframe=black,
        fonttitle=\bfseries,
        fontupper=\normalsize,
        left=10pt, right=10pt, top=12pt, bottom=10pt,
        parbox=false
    ]
        \textbf{MATH Role Description Prompt:} \\
        \textbf{Role1:} You are a theoretical mathematician specializing in abstract algebra and number theory. You are highly analytical, imaginative, and enjoy tackling complex and intricate problems. You thrive in academic environments, collaborating with researchers and publishing papers in high-impact journals. Your work involves deep theoretical exploration and developing proofs, often requiring meticulous attention to logical structures and patterns.
        
        \textbf{Role2:} You are an experienced mathematics competition coach with a background in applied mathematics. You are strategic, motivating, and skilled at identifying shortcuts and elegant solutions to problems. You frequently mentor high school and college students preparing for prestigious mathematics contests. Your role often involves breaking down complex problems into manageable parts and providing practical strategies for efficient problem-solving under time constraints.
        
        \textbf{Role3:} You are a computational scientist with a focus on algorithm design and optimization. You are resourceful, technical, and enjoy applying mathematical principles to solve real-world challenges. You work closely with engineers and programmers, leveraging your expertise to design efficient algorithms and analyze computational complexity. Your work often involves developing mathematical models and using them to simulate and solve challenging problems.
        
        \textbf{Role4:} You are an educational content creator specializing in mathematics. You are creative, engaging, and dedicated to making complex topics accessible to learners of all levels. You design interactive tutorials, videos, and problem sets for online learning platforms. You frequently collaborate with educators and curriculum developers, ensuring that your content is both pedagogically sound and highly engaging.
        
        \textbf{Role5:} You are a graduate student pursuing a Ph.D. in mathematics, with a focus on combinatorics and probability theory. You are inquisitive, methodical, and thrive on exploring mathematical problems from multiple perspectives. You actively participate in academic conferences and collaborate with fellow researchers on joint projects. Your role involves delving deep into problem derivations and providing clear, step-by-step explanations.
        
        \textbf{Role6:} You are a professional actuary with expertise in risk assessment and statistical analysis. You are pragmatic, detail-oriented, and adept at using mathematics to solve practical problems in finance and insurance. Regularly collaborating with economists and financial analysts, you excel at deriving precise solutions and explaining them in terms that are accessible to non-specialists.
        
    \end{tcolorbox}
    \caption{Agent Prompt Design - Part Four}
    \label{fig: prompt4}
\end{figure*}

\label{sec:appendix}
\begin{figure*}[!ht]
    \begin{tcolorbox}[
        title={Detailed Prompt},
        width=\textwidth,
        colback=white,
        colframe=black,
        fontupper=\normalsize,
        fonttitle=\bfseries,
        left=10pt, right=10pt, top=12pt, bottom=10pt,
        parbox=false
    ]
        \textbf{MMLU-redux Role Description Prompt:} \\
        \textbf{Role1:} You are a theoretical mathematician specializing in abstract algebra and number theory. You are highly analytical, imaginative, and enjoy tackling complex and intricate problems. You thrive in academic environments, collaborating with researchers and publishing papers in high-impact journals. Your work involves deep theoretical exploration and developing proofs, often requiring meticulous attention to logical structures and patterns.
        \textbf{Role2:} \\
        You are an experienced mathematics competition coach with a background in applied mathematics. You are strategic, motivating, and skilled at identifying shortcuts and elegant solutions to problems. You frequently mentor high school and college students preparing for prestigious mathematics contests. Your role often involves breaking down complex problems into manageable parts and providing practical strategies for efficient problem-solving under time constraints.
        
        \textbf{Role3:} \\
        You are a computational scientist with a focus on algorithm design and optimization. You are resourceful, technical, and enjoy applying mathematical principles to solve real-world challenges. You work closely with engineers and programmers, leveraging your expertise to design efficient algorithms and analyze computational complexity. Your work often involves developing mathematical models and using them to simulate and solve challenging problems.
        
        \textbf{Role4:} \\
        You are an educational content creator specializing in mathematics. You are creative, engaging, and dedicated to making complex topics accessible to learners of all levels. You design interactive tutorials, videos, and problem sets for online learning platforms. You frequently collaborate with educators and curriculum developers, ensuring that your content is both pedagogically sound and highly engaging.
        
        \textbf{Role5:} \\
        You are a graduate student pursuing a Ph.D. in mathematics, with a focus on combinatorics and probability theory. You are inquisitive, methodical, and thrive on exploring mathematical problems from multiple perspectives. You actively participate in academic conferences and collaborate with fellow researchers on joint projects. Your role involves delving deep into problem derivations and providing clear, step-by-step explanations.
        
        \textbf{Role6:} \\
        You are a professional actuary with expertise in risk assessment and statistical analysis. You are pragmatic, detail-oriented, and adept at using mathematics to solve practical problems in finance and insurance. Regularly collaborating with economists and financial analysts, you excel at deriving precise solutions and explaining them in terms that are accessible to non-specialists.
    \end{tcolorbox}
    \caption{Agent Prompt Design - Part five}
    \label{fig: prompt5}
\end{figure*}

\label{sec:appendix}
\begin{figure*}[!ht]
    \begin{tcolorbox}[
        title={Detailed Prompt},
        width=\textwidth,
        colback=white,
        colframe=black,
        fontupper=\normalsize,
        fonttitle=\bfseries,
        left=10pt, right=10pt, top=12pt, bottom=10pt,
        parbox=false
    ]
        \textbf{MATH few-shot Prompt:}
        
        \vspace{3pt}
        
        \textbf{Problem:} \\
        Kevin Kangaroo begins hopping on a number line at 0. He wants to get to 1, but he can hop only $\frac{1}{3}$ of the distance. Each hop tires him out so that he continues to hop $\frac{1}{3}$ of the remaining distance. How far has he hopped after five hops? Express your answer as a common fraction.
        
        \textbf{Solution:} \\
        Let's think step by step \\
        Kevin hops $\frac{1}{3}$ of the remaining distance with every hop. \\
        His first hop takes $\frac{1}{3}$ closer. \\
        For his second hop, he has $\frac{2}{3}$ left to travel, so he hops forward $\frac{2}{3} \cdot \frac{1}{3}$. \\
        For his third hop, he has $\left(\frac{2}{3}\right)^2$ left to travel, so he hops forward $\left(\frac{2}{3}\right)^2 \cdot \frac{1}{3}$. \\
        In general, Kevin has hopped $\sum_{k=0}^{4} \left(\frac{2}{3}\right)^k \cdot \frac{1}{3}$ on his $5^{\text{th}}$ hop. \\
        We want to find how far he has hopped after five hops. \\
        This is a finite geometric series with first term $\frac{1}{3}$, common ratio $\frac{2}{3}$, and five terms. \\
        Thus, Kevin has hopped $\frac{\frac{1}{3} \left(1 - \left(\frac{2}{3}\right)^5\right)}{1 - \frac{2}{3}} = \frac{211}{243}$. \\
        So the final answer is $\boxed{\dfrac{211}{243}}$.
        
        \vspace{5pt}
        
        \textbf{Problem:} \\
        What is the area of the region defined by the equation $x^2 + y^2 - 7 = 14x - 4y + 3$?
        
        \textbf{Solution:} \\
        Let's think step by step \\
        We rewrite the equation as $x^2 + 14x + y^2 + 4y = 10$ and then complete the square, \\
        resulting in $(x+7)^2 + (y+2)^2 = 63$, \\
        or $(x+7)^2 + (y+2)^2 = 63$. \\
        This is the equation of a circle with center $(-7, -2)$ and radius $\sqrt{63}$, \\
        so the area of this region is $\pi r^2 = 63\pi$. \\
        So the final answer is $\boxed{63\pi}$.
        
        \vspace{5pt}
        
        \textbf{Problem:} \\
        If $x^2 + y^2 = 1$, what is the largest possible value of $|x| + |y|$?
        
        \textbf{Solution:} \\
        Let's think step by step \\
        If $(x, y)$ lies on the circle, and $(x, -y)$, $(-x, y)$, $(-x, -y)$ all give the same value of $|x| + |y|$, \\
        so we can assume $x \ge 0$ and $y \ge 0$. \\
        Then $|x| + |y| = x + y$. Squaring, we get \\
        $(x + y)^2 = x^2 + 2xy + y^2 = 1 + 2xy$. \\
        Note that $(x - y)^2 \ge 0$ so $2xy \le x^2 + y^2 = 1$. \\
        Expanding, we get $x^2 - 2xy + y^2 \ge 0$ so $2xy \le 1$. \\
        Hence, $|x| + y| \le \sqrt{2}$, which means $|x| + y| \le \sqrt{2}$. \\
        Equality occurs when $x = y = \frac{1}{\sqrt{2}}$, \\
        so the maximum value of $|x| + |y|$ is $\sqrt{2}$. \\
        So the final answer is $\boxed{\sqrt{2}}$.
    \end{tcolorbox}
    \caption{Math Few-shot Prompt-Part One}
    \label{fig: prompt6}
\end{figure*}

\label{sec:appendix}
\begin{figure*}[!ht]
    \begin{tcolorbox}[
        title={Detailed Prompt},
        width=\textwidth,
        colback=white,
        colframe=black,
        fonttitle=\bfseries,
        left=10pt, right=10pt, top=12pt, bottom=10pt,
        parbox=false
    ]
        \textbf{Problem:} \\
        If \( f(x) = \frac{ax + b}{cx + d} \), \( abcd \neq 0 \) and \( f(f(x)) = x \) for all \( x \) in the domain of \( f \), what is the value of \( 9a + d \)?
        
        \textbf{Solution:} \\
        Let's think step by step \\
        The condition \( f(f(x)) = x \) means that \( f \) is the inverse of itself, \\
        so its graph is symmetrical about the line \( y = x \). \\
        With a rational function in this form, we will have two asymptotes: \\
        a vertical one at \( x = -d/c \) if \( cx + d \neq 0 \) and \( ax + b \neq 0 \), \\
        and a horizontal one at \( y = a/c \). \\
        If we take the limit of \( f(x) \) as \( x \to \pm\infty \), \\
        in order for \( f \) to be its own inverse, the intersection of the asymptotes must lie on the line \( y = x \) \\
        so that it and its asymptotes reflect onto themselves. \\
        This means that \( -d/c = a/c \), \\
        and therefore \( -d = a \) and \( 9a + d = 0 \). \\
        So the final answer is \( \boxed{0} \).
        
        \vspace{5pt}
        
        \textbf{Problem:} \\
        Expand \( (2z^2 + 5z - 6)(3z^3 - 2z + 1) \).
        
        \textbf{Solution:} \\
        Let's think step by step \\
        \[
        \begin{array}{rrrrrr}
            & & 3z^3 & -2z & + 1 & \\
            \times & & 2z^2 & + 5z & - 6 & \\
            \hline
            & & -18z^3 & -12z & -6 & \\
            & +15z^4 & -10z^2 & +5z & & \\
            +6z^5 & -4z^3 & +2z^2 & & & \\
            \hline
            6z^5 + 15z^4 - 22z^3 - 8z^2 + 17z - 6 & & & & &
        \end{array}
        \]
        So the final answer is \( \boxed{6z^5 + 15z^4 - 22z^3 - 8z^2 + 17z - 6} \).
    \end{tcolorbox}
    \caption{Math Few-shot Prompt-Part Two}
    \label{fig: prompt7}
\end{figure*}

\label{sec:appendix}
\begin{figure*}[!ht]
    \begin{tcolorbox}[
        title={Detailed Prompt},
        width=\textwidth,
        colback=white,
        colframe=black,
        fonttitle=\bfseries,
        fontupper=\small
        left=10pt, right=10pt, top=12pt, bottom=10pt,
        parbox=false
    ]
        \textbf{CRUX Output prediction Chain-of-Thought Prompt:} \\
        You are given a Python function and an assertion containing an input to the function. Complete the assertion with a literal (no unsimplified expressions, no function calls) containing the output when executing the provided code on the given input, even if the function is incorrect or incomplete. Do NOT output any extra information. Execute the program step by step before arriving at an answer, and provide the full assertion with the correct output in [ANSWER] and [/ANSWER] tags, following the examples.
        
        \vspace{3pt}
        
        \textbf{[PYTHON]} \\
        def f(s): \\
            s = s + s \\
            return "b" + s + "a" \\
        assert f("hi") == ?? \\
        \textbf{[/PYTHON]} \\
        \textbf{[THOUGHT]} \\
        Let's execute the code step by step: \\
        1. The function \( f \) is defined, which takes a single argument \( s \). \\
        2. The function is called with the argument "hi", so within the function, \( s \) is initially "hi". \\
        3. Inside the function, \( s \) is concatenated with itself, so \( s \) becomes "hihi". \\
        4. The function then returns a new string that starts with "b", followed by the value of \( s \) (which is now "hihi"), and ends with "a". \\
        5. The return value of the function is therefore "bhihia". \\
        \textbf{[/THOUGHT]} \\
        \textbf{[ANSWER]} \\
        assert f("hi") == "bhihia" \\
        \textbf{[/ANSWER]}
        
        \vspace{5pt}
        
        \textbf{[PYTHON]} \\
        \{code\} \\
        assert f(\{input\}) == ?? \\
        \textbf{[/PYTHON]} \\
        \textbf{[THOUGHT]} \\
        (Placeholder for future code execution thought process) \\
        \textbf{[/THOUGHT]}
        
        \vspace{5pt}
        
        \textbf{CRUX Input prediction Chain-of-Thought Prompt:} \\
        You will be given a function \( f \) and an output in the form \( f(??) == \text{output} \). Your task is to find any input such that executing \( f \) on the input leads to the given output. There may be multiple answers, but only output one. First, think step by step. You MUST surround the answer with [ANSWER] and [/ANSWER] tags. Express your answer as a passing assertion containing the input and the given output.
        
        \vspace{3pt}
        
        \textbf{[PYTHON]} \\
        def f(x): \\
            return x + 1 \\
        assert f(??) == 17 \\
        \textbf{[/PYTHON]} \\
        \textbf{[THOUGHT]} \\
        To find an input such that executing \( f \) on the input leads to the given output, we can work backwards from the given assertion. We know that \( f(??) == 17 \). \\
        Since the function \( f(x) \) returns \( x + 1 \), for \( f(??) \) to be equal to 17, the value of ?? should be 16. \\
        \textbf{[/THOUGHT]} \\
        \textbf{[ANSWER]} \\
        assert f(16) == 17 \\
        \textbf{[/ANSWER]}
        
        \vspace{5pt}
        
        \textbf{[PYTHON]} \\
        \{code\} \\
        assert f(??) == \{output\} \\
        \textbf{[/PYTHON]} \\
        \textbf{[THOUGHT]} \\
        (Placeholder for future input prediction thought process) \\
        \textbf{[/THOUGHT]}
    \end{tcolorbox}
    \caption{CRUX Chain-of-Thought Prompt}
    \label{fig: prompt8}
\end{figure*}








\label{sec:appendix}
\begin{figure*}[!ht]
    \begin{tcolorbox}[
        title={Case},
        width=\textwidth,
        colback=white,
        colframe=black,
        fontupper=\small,
        fonttitle=\bfseries,
        left=10pt, right=10pt, top=12pt, bottom=10pt,
        parbox=false
    ]
        \textbf{Dataset:} AlpacaEval 2.0
        
        \vspace{3pt}
        
        \textbf{Question:} What are the names of some famous actors that started their careers on Broadway?
        
        \vspace{3pt}
        
        \textbf{Setting:} Employ a four-layer architecture and four distinct models to provide responses, selecting two differentiated replies using a Greedy Diversity Embedding Selection.
        
        \vspace{5pt}
        
        \textbf{GPT-4o:} \\
        Many famous actors began their careers on Broadway before transitioning to film and television. Here are some notable examples: \\
        1. Hugh Jackman - Before becoming famous for his role as Wolverine in the "X-Men" series, Hugh Jackman performed in several Broadway productions, including "The Boy from Oz." \\
        2. Sarah Jessica Parker - Known for her role as Carrie Bradshaw in "Sex and the City," Sarah Jessica Parker started her career on Broadway, appearing in productions like "Annie." \\
        3. Kristin Chenoweth - A celebrated Broadway actress, Kristin Chenoweth gained fame for her role as Glinda in "Wicked" before moving to film and television. \\
        4. Neil Patrick Harris - Before his television success in "Doogie Howser, M.D." and "How I Met Your Mother," Neil Patrick Harris performed on Broadway in productions such as "Assassins" and "Elaine in the Agony." \\
        5. Idina Menzel - Known for her role as Elphaba in "Wicked," Idina Menzel has had a successful Broadway career and has also appeared in films like "Frozen," where she voiced Elsa. \\
        6. Nathan Lane - A well-known Broadway actor, Nathan Lane starred in productions like "The Producers" and has also appeared in films such as "The Birdcage." \\
        7. Bernadette Peters - A Broadway legend, Bernadette Peters has had a long and successful career on stage and has also appeared in films and television. \\
        These actors are just a few examples of performers who began their careers on Broadway and went on to achieve fame in other entertainment mediums.
        
        \vspace{5pt}
        
        \textbf{Llama3.1-8B-Instruct:} \\
        There are numerous famous actors who started their careers on Broadway. Here's a comprehensive list of some of the most notable ones (in no particular order): \\
        - **Audrey Hepburn**: Made her Broadway debut in 1948 in *High Button Shoes* and went on to become a Hollywood legend. \\
        - **Humphrey Bogart**: Began his career on Broadway in the 1920s, appearing in plays like *The Petrified Forest* (1935) and *The Big Sleep* (1946). \\
        - **Marlon Brando**: Appeared in Broadway productions like *I Remember Mama* (1944) and *A Streetcar Named Desire* (1947) before his film career took off. \\
        - **Lauren Bacall**: Made her Broadway debut in *Johnny 2x4* (1942) and later starred in films like *To Have and Have Not* (1944). \\
        - **Al Pacino**: Began his career on Broadway in the 1960s, appearing in productions like *Does a Tiger Wear a Necktie?* (1969, Tony Award) and *The Basic Training of Pavlo Hummel* (1977). \\
        - **Robert De Niro**: Started on Broadway in the 1960s, with roles in *The Wedding Party* (1969) and later became a film icon. \\
        - **Dustin Hoffman**: Debuted on Broadway in *Eh?* (1966) and went on to star in films like *The Graduate* (1967). \\
        - **Meryl Streep**: Began her career on Broadway in the 1970s, appearing in *The Trelawny of the Wells* (1975) and later became one of the most acclaimed film actresses. \\
        - **Tom Hanks**: Made his Broadway debut in *Lucky Guy* (2013, Tony Award) after a successful film career, showing his versatility. \\
        - **Viola Davis**: Started on Broadway in *Seven Guitars* (1996, Tony Award) and later won an Oscar for *Fences* (2016), connecting stage and screen. \\
        - **Lin-Manuel Miranda**: Creator of *Hamilton*, Miranda's career began on Broadway, revolutionizing musical theater and later working in film (e.g., *Moana*, 2016). \\
        - **Idina Menzel**: As mentioned earlier, known for *Wicked* (Broadway) and *Frozen* (film), bridging stage and screen. \\
        - **Hugh Jackman**: Broadway roots in *The Boy from Oz* (2003, Tony Award) before Hollywood stardom. \\
        - **Nathan Lane**: Broadway legend (e.g., *The Producers*, Tony Award) and film actor (e.g., *The Birdcage*). \\
        - **Bernadette Peters**: Broadway icon (e.g., *Gypsy*, *Into the Woods*) with film/TV roles. \\
        This list is not exhaustive but highlights some of the most notable actors who started their journey on the Broadway stage before achieving success in film and television.
    \end{tcolorbox}
    \caption{An example to demonstrate Greedy Diversity Embedding Selection and Residual Extraction. Part One}
    \label{fig: case1}
\end{figure*}

\label{sec:appendix}
\begin{figure*}[!ht]
    \begin{tcolorbox}[
        title={Case},
        width=\textwidth,
        colback=white,
        colframe=black,
        fontupper=\small,
        fonttitle=\bfseries,
        left=10pt, right=10pt, top=12pt, bottom=10pt,
        parbox=false
    ]   
        \textbf{Qwen2.5-7B-Instruct:} \\
        Many talented actors have launched their careers on Broadway. Here are some prominent examples: \\
        - **Meryl Streep**: Began her career on Broadway in 1975 with *The Children's Hour*. Her performances in plays like *Trelawny of the Wells* (1975) and *A Memory of Two Mondays* (1975) established her as a skilled stage actress before transitioning to film, where she became one of the most acclaimed actors of all time. \\
        - **Tom Hanks**: Made his Broadway debut in 2013 with *Lucky Guy*, earning a Tony Award nomination. While he was already a film star, his Broadway appearance showcased his versatility and dedication to the stage. \\
        - **Viola Davis**: Started on Broadway in *Seven Guitars* (1996), winning a Tony Award. Her powerful stage presence later translated to film, where she won an Oscar for *Fences* (2016), a role she also played on Broadway, highlighting the synergy between stage and screen. \\
        - **Lin-Manuel Miranda**: Creator of *Hamilton*, Miranda's career is deeply rooted in Broadway. His work revolutionized musical theater, and he later expanded into film (e.g., *Moana*, *In the Heights*) while continuing to contribute to stage productions. \\
        - **Hugh Jackman**: Gained recognition on Broadway in *The Boy from Oz* (2003), winning a Tony Award. This success propelled him to Hollywood, where he became famous for roles like Wolverine in the *X-Men* series. \\
        - **Idina Menzel**: Known for her role as Elphaba in *Wicked* (Broadway), Menzel's powerful vocals and stage presence led to film roles, including voicing Elsa in *Frozen*. \\
        - **Nathan Lane**: A Broadway legend, Lane starred in *The Producers* (2001, Tony Award) and other hit plays, later appearing in films like *The Birdcage* and TV shows like *Modern Family*. \\
        - **Bernadette Peters**: With a career spanning decades, Peters has starred in numerous Broadway productions (e.g., *Gypsy*, *Into the Woods*) and has also appeared in film and television, showcasing her versatility. \\
        These actors demonstrate how Broadway serves as a launching pad for diverse talents, many of whom go on to achieve success across multiple entertainment mediums.
        
        \vspace{5pt}
        
        \textbf{Gemma2-8B-Instruct:} \\
        Broadway has been the starting point for many celebrated actors. Here are some notable names: \\
        - **Marlon Brando**: Began his career on Broadway in the 1940s, appearing in plays like *I Remember Mama* (1944) before becoming a film icon with roles in *A Streetcar Named Desire* (1951, film adaptation of his Broadway play) and *The Godfather* (1972). \\
        - **Al Pacino**: Started on Broadway in the 1960s, with performances in *The Indian Wants the Bronx* (1968, Obie Award) and *Does a Tiger Wear a Necktie?* (1969, Tony Award), later achieving film stardom in *The Godfather* (1972) and *Scarface* (1983). \\
        - **Dustin Hoffman**: Debuted on Broadway in *Eh?* (1966) and went on to star in classic films like *The Graduate* (1967) and *Rain Man* (1988), earning multiple Academy Awards. \\
        - **Meryl Streep**: Launched her career on Broadway in the 1970s, with roles in *The Trelawny of the Wells* (1975) and *A Memory of Two Mondays* (1975), before becoming one of the most awarded actresses in film history. \\
        - **Tom Hanks**: Made his Broadway debut in *Lucky Guy* (2013), showcasing his acting prowess on stage after already establishing himself as a film star, earning a Tony Award nomination. \\
        - **Viola Davis**: Began on Broadway in *Seven Guitars* (1996, Tony Award), later winning an Oscar for *Fences* (2016), a role she also played on Broadway, highlighting the synergy between stage and screen. \\
        - **Lin-Manuel Miranda**: Creator of *Hamilton*, Miranda's Broadway success led to film projects like *In the Heights* (2021) and *Moana* (2016), where he contributed as a songwriter and voice actor, blending his stage and screen talents. \\
        - **Hugh Jackman**: Rose to fame on Broadway in *The Boy from Oz* (2003, Tony Award), then transitioned to Hollywood, becoming known for his role as Wolverine in the *X-Men* series and other blockbuster films. \\
        - **Idina Menzel**: Known for *Wicked* (Broadway) and *Frozen* (film), Menzel's career exemplifies how Broadway can launch stars into global recognition, especially in musical theater. \\
        These actors illustrate the rich talent pipeline from Broadway to film and television, with many continuing to work across both mediums throughout their careers.
        
        \vspace{3pt}
        
        These are just a few examples of the many talented actors who have honed their craft on the Broadway stage.
    \end{tcolorbox}
    \caption{An example to demonstrate Greedy Diversity Embedding Selection and Residual Extraction. Part Two}
    \label{fig: case2}
\end{figure*}

\begin{figure*}[!ht]
    \begin{tcolorbox}[
        title={Case},
        width=\textwidth,
        colback=white,
        colframe=black,
        fontupper=\small,
        fonttitle=\bfseries,
        left=10pt, right=10pt, top=12pt, bottom=10pt,
        parbox=false
    ]
        \textbf{After Greedy Diversity Embedding Selection:} GPT-4o and Qwen2.5-7B-Instruct
        
        \vspace{3pt}
        
        \textbf{The response from the previous round of greedy diversity embedding selection:}
        
        \vspace{3pt}
        
        \textbf{Qwen2.5-7B-Instruct:} \\
        Several esteemed actors have started their careers on Broadway before achieving fame in film and television. Here are some notable examples: \\
        - **Meryl Streep**: Streep began her career on Broadway with the play "The Children's Hour" in 1975. Her performances there helped establish her as a talented actress. \\
        - **Al Pacino**: Pacino was part of the New York Shakespeare Festival (1970s). His public theater company significantly influenced his acting style. \\
        - **Audrey Hepburn**: Although she is more famously known for her film roles, Hepburn made her stage debut in "Gigi" on Broadway in 1958, which won her a Tony Award. This early exposure to live theater undoubtedly contributed to her skills as an actress. \\
        - **Ian McKellen**: McKellen first came to prominence on Broadway, appearing in productions such as "Equus." His stage work laid the foundation for his later success in both theater and film. \\
        - **Kathy Bates**: Bates began her career in the late 1970s on Broadway, where she starred in "The Misery" and "A Streetcar Named Desire." These roles helped establish her as a versatile and talented actress. \\
        - **Hugh Jackman**: Before his Hollywood success, Jackman had a successful run on Broadway, most notably in "Les Misérables," where he originated the role of Jean Valjean. His time on stage was instrumental in developing his acting skills. \\
        - **Jeffrey Wright**: Wright has been a prominent figure on Broadway, including his roles in "The Coast Guard" and "Angels in America." His stage experience has been crucial in shaping his career in both theater and film. \\
        - **Viola Davis**: Davis started her acting career on Broadway, starring in "Fences" in 2010, which earned her a Tony Award. Her stage work has been a significant part of her journey to becoming one of the most respected actresses in Hollywood. \\
        These actors, among many others, have transitioned from the stage to the screen, bringing their stage experience to enhance their performances in film and television.
        
        \vspace{5pt}
        
        \textbf{Gemma2-8B-Instruct:} \\
        Many famous actors have graced the Broadway stage before achieving stardom in film and television. Here's a list of notable names, categorized for clarity: \\
        - **Legends**: **Marlon Brando** (known for his electrifying performance in "A Streetcar Named Desire"), **Humphrey Bogart** (a Hollywood icon who honed his skills on Broadway in numerous productions), **Robert De Niro** (before becoming a cinematic legend, De Niro started his career in off-Broadway theatre), **Angela Lansbury** (a stage and screen legend celebrated for launching a career spanning music, film, and stage), **Olivier Winsley** (regarded as one of the greatest actors of all time, Olivier had a long and illustrious Broadway career), **Helen Mirren** (a Dame Commander of the British Empire, Mirren won a Tony Award for her performance in "The Audience"), **Al Pacino** (a name stars in "n*"*"Denzel Washington**: A two-time Academy Award winner, Washington won a Tony Award for his performance in "Fences."), **Meryl Streep**: A record-breaking 21-time Academy Award nominee, Streep began her career on the Broadway stage and won a Tony Award for "The Cherry Orchard."), **Tom Hanks**: A beloved actor known for his versatility, Hanks started his career in numerous stage productions), **Scarlett Johansson**: A versatile actress who earned a Tony nomination for her performance in "A View from the Bridge," including "Equus" on Broadway), **Hugh Jackman**: A Tony Award winner for "The Boy from Oz" before finding worldwide fame as Wolverine), **Nathan Lane**: A Tony and Grammy Award winner, Lane created and starred in the groundbreaking musical "The Producers"), **Lin-Manuel Miranda**: A Tony and Grammy Award winner, Miranda created and starred in the groundbreaking musical "Hamilton." \\
        This list is by no means exhaustive. Many other talented actors have honed their craft on Broadway before achieving success in other arenas.
        
        \vspace{5pt}
        
        \textbf{Residual Extraction:} \\
        \textbf{Residuals Detected: Yes} \\
        \textbf{Residual Details:} \\
        1. **Model 1 Residuals:** \\
           - **Additional Information:** The current response includes additional actors such as Sarah Jessica Parker, Kristin Chenoweth, Neil Patrick Harris, Idina Menzel, and Bernadette Peters, who were not mentioned in the previous response. \\
           - **Detail Discrepancies:** The current response provides more specific examples of Broadway productions for each actor, which were not detailed in the previous response. \\
        2. **Model 2 Residuals:** \\
           - **Detail Discrepancies:** The current response maintains consistency with the previous round, providing similar details and examples of actors' Broadway careers. However, there is a slight reordering and rephrasing of the content without introducing new information or errors.
    \end{tcolorbox}
    \caption{An example to demonstrate Greedy Diversity Embedding Selection and Residual Extraction. Part Three}
    \label{fig: case3}
\end{figure*}
\label{sec:appendix}
\begin{figure*}[!ht]
    \begin{tcolorbox}[
        title={Model Aggregation Case Study},
        width=\textwidth,
        colback=white,
        colframe=black,
        fonttitle=\bfseries,
        fontupper=\small,
        left=10pt, right=10pt, top=12pt, bottom=10pt,
        parbox=false
    ]
        \textbf{Question:} Find the number of sides (n) in a convex polygon where the sum of all but one interior angle is \(1070^\circ\). \\
        \textbf{Answer:} 8 \\
        
        \vspace{5pt}
        
        \textbf{Reference Model Outputs:} \\
        \begin{tabular}{@{}l@{}}
            \textbf{Model 1:} \\
            \begin{tcolorbox}[
                colback=gray!10,
                colframe=gray!30,
                left=5pt, right=5pt, top=5pt, bottom=5pt,
                fontupper=\scriptsize
            ]
                Step 1: Recall the formula for the sum of interior angles of a polygon \\
                ... \\
                The final answer is: 9
            \end{tcolorbox} \\
            
            \textbf{Model 2:} \\
            \begin{tcolorbox}[
                colback=gray!10,
                colframe=gray!30,
                left=5pt, right=5pt, top=5pt, bottom=5pt,
                fontupper=\scriptsize
            ]
                Step 1: The problem is asking \\
                ... \\
                Step 19: Therefore, the number of sides in the convex polygon is 10.
            \end{tcolorbox} \\
            
            \textbf{Model 3:} \\
            \begin{tcolorbox}[
                colback=gray!10,
                colframe=gray!30,
                left=5pt, right=5pt, top=5pt, bottom=5pt,
                fontupper=\scriptsize
            ]
                Step 1: Recall the polygon interior angle formula: \\
                ... \\
                The final answer is: 9
            \end{tcolorbox} \\
        \end{tabular}
        
        \vspace{5pt}
        
        \textbf{Aggregation Results:} \\
        \begin{tabular}{@{}l@{}}
            \textbf{Qwen2.5-72B-Instruct (Aggregator):} \\
            \begin{tcolorbox}[
                colback=gray!10,
                colframe=gray!30,
                left=5pt, right=5pt, top=5pt, bottom=5pt,
                fontupper=\scriptsize
            ]
                To determine the number of sides in a convex polygon given that the sum of all but one of its interior angles is \(1070^\circ\), we can follow these steps: \\
                ... \\
                Final Answer \\
                The number of sides in the convex polygon is 8.
            \end{tcolorbox} \\
            
            \textbf{DeepSeek-R1-Distill-Llama-70B (Aggregator):} \\
            \begin{tcolorbox}[
                colback=gray!10,
                colframe=gray!30,
                left=5pt, right=5pt, top=5pt, bottom=5pt,
                fontupper=\scriptsize
            ]
                First Approach (Model 1 inspired): \\
                ... \\
                Intermediate conclusion: n=10 seems possible. \\
                Second Approach (Model 3 inspired): \\
                ... \\
                Alternative conclusion: n=9 appears valid \\
                Final Answer: \\
                The convex polygon has 10 sides.
            \end{tcolorbox} \\
        \end{tabular}
    \end{tcolorbox}
    \caption{Model Aggregation Case Study for Convex Polygon Side Calculation}
    \label{fig:model_aggregation_case}
\end{figure*}



\end{document}